%% file: main.tex
\documentclass[sigconf,authorversion,nonacm]{acmart}

\AtBeginDocument{%
  \providecommand\BibTeX{{%
    \normalfont B\kern-0.5em{\scshape i\kern-0.25em b}\kern-0.8em\TeX}}}

\usepackage{tikz}
\usepackage{makecell}

\usepackage{amsmath,amssymb,amsfonts,bbm}
\usepackage{enumitem}
\usepackage{algorithm}
\usepackage{algpseudocode}
\usepackage{diagbox}
\usepackage{graphicx}

\usepackage{textcomp}
\usepackage{xcolor,xspace}
\usepackage{mathrsfs}
\usepackage{booktabs}
\usepackage{wasysym}
\usepackage{pifont}
\usepackage{filecontents}
\usepackage{threeparttable}
\usepackage{dsfont}
\input{prestuff}

\newcommand{\system}{\textsc{MeBench}\xspace}

\newcommand{\facepp}{Face$^{++}$\xspace}

\begin{document}
%-------------------------------------------------------------------------------

%don't want date printed
\date{}

\title{Model Extraction Attacks Revisited}
\author{Jiacheng Liang}
\affiliation{%
  \institution{Stony Brook University}
  \city{New York}
  \country{USA}}
\email{ljcpro@outlook.com}

\author{Ren Pang}
\affiliation{%
  \institution{Penn State University}
  \city{State College}
  \country{USA}}
\email{rbp5354@psu.edu}

\author{Changjiang Li}
\affiliation{%
  \institution{Stony Brook University}
  \city{New York}
  \country{USA}}
\email{meet.cjli@gmail.com}

\author{Ting Wang}
\affiliation{%
  \institution{Stony Brook University}
  \city{New York}
  \country{USA}}
\email{inbox.ting@gmail.com}

\renewcommand{\shortauthors}{Liang and Pang, et al.}
\input{abstract}





% \thispagestyle{plain}
% \pagenumbering{arabic}
% \pagestyle{plain}

\maketitle

\input{Introduction}

\input{Fundamentals}

\input{Threat_Model}
\input{Experiment}
\input{literature}
\input{Conclusion}
\newpage

\bibliographystyle{ACM-Reference-Format}
\bibliography{cite}
\input{appendix.tex}
\end{document}

%% file: prestuff.tex
\usepackage{epsfig,amsmath,amsfonts,epsfig,multirow,makecell,caption,soul,csquotes,color,wrapfig,subcaption,mathtools,bm,spverbatim,booktabs,tcolorbox,diagbox}
\usepackage[e]{esvect}

\usepackage{caption}
\makeatletter
\newif\if@restonecol
\makeatother
\newenvironment{changemargin}[2]{\begin{list}{}{
	\setlength{\topsep}{0pt}\setlength{\leftmargin}{0pt}
	\setlength{\rightmargin}{0pt}
	\setlength{\listparindent}{\parindent}
	\setlength{\itemindent}{\parindent}
	\setlength{\parsep}{0pt plus 1pt}
	\addtolength{\leftmargin}{#1}\addtolength{\rightmargin}{#2}
	}\item}
	{\end{list}}

\usepackage[first=0,last=9]{lcg}
\usepackage{colortbl}
\definecolor{Gray}{gray}{0.8}
\colorlet{Red}{red!10!white}
\colorlet{Blue}{blue!10!white}

\usepackage{hyperref}

\newcommand{\msec}[1]{\S\ref{#1}}
\newcommand{\mref}[1]{\,\ref{#1}}

\newcommand{\mcite}[1]{\,\cite{#1}}

\newcommand{\meg}{\textit{e.g.}\xspace}
\newcommand{\mie}{\textit{i.e.}\xspace}

\newcommand{\mct}[1]{({\it #1})}

\newtcolorbox{mtbox}[1]{left=0.25mm, right=0.25mm, top=0.25mm, bottom=0.25mm, sharp corners, colframe=red!50!black, boxrule=0.5pt, title={#1}, fonttitle=\bfseries, coltitle=red!50!black, attach title to upper={\ --\ }}

\usepackage{scalerel}[2016/12/29]

\makeatletter
\providecommand{\leadsfrom}{%
  \mathrel{\mathpalette\reflect@squig\relax}%
}
\newcommand{\reflect@squig}[2]{%
  \reflectbox{$\m@th#1\leadsto$}%
}
\makeatother

\def\eqref#1{equation~\ref{#1}}
\def\1{\bm{1}}

\DeclareMathAlphabet{\mathsfit}{\encodingdefault}{\sfdefault}{m}{sl}
\SetMathAlphabet{\mathsfit}{bold}{\encodingdefault}{\sfdefault}{bx}{n}

\def\gA{{\mathcal{A}}}

\def\gD{{\mathcal{D}}}

\def\gU{{\mathcal{U}}}

\def\gX{{\mathcal{X}}}
\def\gY{{\mathcal{Y}}}

\newcommand{\E}{\mathbb{E}}

\newcommand{\R}{\mathbb{R}}

%% file: abstract.tex
\begin{abstract}
Model extraction (ME) attacks represent one major threat to Machine-Learning-as-a-Service (MLaaS) platforms by ``stealing'' the functionality of confidential machine-learning models through querying black-box APIs. Over seven years have passed since ME attacks were first conceptualized in the seminal work\mcite{tramer2016stealing}. During this period, substantial advances have been made in both ME attacks and MLaaS platforms, raising the intriguing question: {\em How has the vulnerability of MLaaS platforms to ME attacks been evolving?}

In this work, we conduct an in-depth study to answer this critical question. Specifically, we characterize the vulnerability of current, mainstream MLaaS platforms to ME attacks from multiple perspectives including attack strategies, learning techniques, surrogate-model design, and benchmark tasks. Many of our findings challenge previously reported results, suggesting emerging patterns of ME vulnerability.
Further, by analyzing the vulnerability of the same MLaaS platforms using historical datasets from the past four years, we retrospectively characterize the evolution of ME vulnerability over time, leading to a set of interesting findings. Finally, we make suggestions about improving the current practice of MLaaS in terms of attack robustness. Our study sheds light on the current state of ME vulnerability in the wild and points to several promising directions for future research.
\end{abstract}

%% file: Introduction.tex
% \section{Introduction}

% \subsection{Services}

% accuracy, vulnerability, potential defenses (e.g., quantization), transferability

% \subsection{Attacks}

% knowledge distillation/mixmatch/adaptive learning (their potential combination), adam/sgd, quantization (why it works), piracy model (resnet, bit, vit, width versus depth), query budget, transferability, augmentation, data imbalance, transferability of adversarial attack 

% \subsection{Tasks}

% how the attack vulnerabilities differ across different tasks? -- image classification, NLP recognition, 

% \subsection{Times}

\section{Introduction}
% \jiang{The exponential rise of deep learning, marked by breakthroughs in language and generative models, has spurred the creation of a burgeoning industry - Machine Learning as a Service (MLaaS). This sector meets the increasing demand for advanced AI by offering platforms like ChatGPT and Midjourney, enabling businesses and individuals to utilize high-end AI without substantial expertise or infrastructure. However, the high development cost of these models makes them valuable intellectual assets, and the widespread adoption of MLaaS has heightened security risks, notably model extraction attacks.  }

% The popularity of Deep learning has grown rapidly in recent years, with its applications in a lot of domains \mcite{he2016deep,deng2014ensemble,roy2018deep,naumov2019deep,vr,lagler2013gpt2}. However, the cost of building and training these models can be prohibitive for many organizations, leading to the emergence of Machine Learning as a Service (MLaaS)\cite{ribeiro2015mlaas}. 

% MLaaS is a cloud-based service that enables users to access pre-trained ML models or build their own ML models using pre-built tools and libraries. Despite the benefits of MLaaS, there are several security risks associated with this approach. First, MLaaS is the private property of the owner, but it may be subject to theft. 

The remarkable advances in machine learning (ML) technologies in recent years\mcite{zhou2023pass,zhang2021restore,chen2023dark,li2024feature,liang2021omnilytics,qi2023ocbev,xu2023toposemiseg} have spurred the expansion of Machine-Learning-as-a-Service (MLaaS) \mcite{li2022seeing,shi2018active}, which meets the increasing need for ML capabilities among users who might not possess the requisite expertise or infrastructure. MLaaS platforms offer publicly accessible APIs \mcite{li2022seeing,feng2021live}, enabling users to interface with backend ML models and systems seamlessly. Users are often charged on a per-query basis, making advanced ML capabilities both accessible and affordable. Many IT giants (\meg, Google, Amazon, Microsoft, and \facepp) have unveiled their MLaaS platforms.

% advanced AI by offering platforms such as ChatGPT\mcite{chatgpt} and Midjourney\mcite{midjourney}, enabling businesses and individuals to utilize high-end AI without substantial expertise or infrastructure. However, the high development cost of these models makes them valuable intellectual assets, and the widespread adoption of MLaaS has heightened security risks, notably model extraction (ME) attacks. 

\begin{figure}[!ht]
    \centering
    \includegraphics[width=0.27\textwidth]{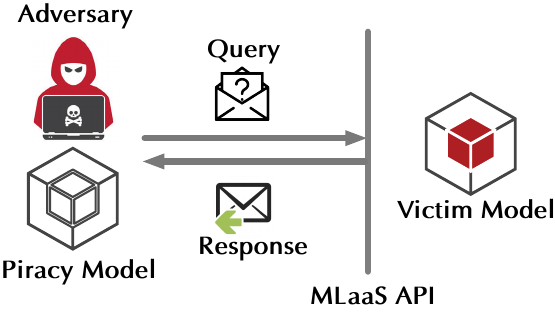}
    \caption{Model extraction attacks.}
    \label{fig:me}
\end{figure}

However, there exists an inherent conflict between the proprietary nature of ML models and the public accessibility of MLaaS APIs, leading to a range of security concerns \mcite{liu2023long,liu2022efficient,li2019det, li2023embarrassingly, li2021towards, liu2024please,liu2023slowlidar,lyu2024task,lyu2023attention,huangadversarial}. One of the most prominent concerns is the threat of model extraction (ME) attacks. As illustrated in Figure\mref{fig:me}, such attacks construct a piracy model functionally equivalent or similar to the backend model of MLaaS, via querying the black-box APIs with carefully crafted inputs. Since its conceptualization in\mcite{tramer2016stealing}, a stream of work has been proposed to improve the efficacy of ME attacks\mcite{tramer2016stealing, orekondy2019knockoff, krishna2019thieves, chandrasekaran2020exploring, pal2019framework, pal2020activethief, pengcheng2018query, shi2018active, juuti2019prada, jagielski2020high, yu2020cloudleak, Truong_2021_CVPR, carlini2021extracting, papernot2016transferability, papernot2016distillation, papernot2018sok, papernot2017practical}. 

Despite the plethora of prior work, our understanding of the vulnerability of real-world MLaaS to ME attacks remains limited. With the exception of \mcite{yu2020cloudleak}, most of these studies simulate ME attacks in controlled environments, which may differ significantly from real-world MLaaS scenarios. For instance, some studies (\meg,\mcite{krishna2019thieves}) assume the adversary possesses prior knowledge about the victim model's architecture,  while others (\meg, \mcite{Truong_2021_CVPR}) create synthetic data to extract models, but MLaaS will initially reject the random noise images, impeding progress. Moreover, since the conceptualization of ME attacks, substantial advances have been made in ME attacks, ML techniques, and MLaaS platforms. It is unclear whether the conclusions drawn in prior work still hold in this rapidly evolving landscape. Specifically, we have the following intriguing questions:

\vspace{1pt}
RQ1 - How vulnerable are today's real-world MLaaS APIs with respect to ME attacks? 

\vspace{1pt}
RQ2 - How might the adversary exploit such vulnerability effectively (\meg, query-cost reduction)? 

\vspace{1pt}
RQ3 - How has the vulnerability of MLaaS platforms been evolving in the past years?

\vspace{2pt}
{\bf Our work --}
To answer these questions, we design, implement, and evaluate \system, the first open-source platform for evaluating the ME vulnerability of MLaaS APIs in a unified and holistic manner. Currently, \system has integrated 4 representative ME attacks, 4 attack performance metrics, as well as a suite of 11 piracy models and 6 benchmark datasets. Further, \system has implemented a rich set of analysis tools for characterizing the vulnerability, including comparing vulnerability across different APIs, measuring attack transferability, and tracing vulnerability evolution.
Our findings can be summarized as follows.

\begin{table*}[!ht]\small
    \setlength\tabcolsep{10pt}
    \renewcommand{\arraystretch}{1.2}
    \centering
    \begin{tabular}{m{6cm}|m{8cm}|c}
    Previous Conclusion &Refined Conclusion  & Consistency\\
    \hline
    \hline
    Identifying the architectures of victim models and using the same architectures in piracy models substantially boost the effectiveness of ME attacks\mcite{chen2022teacher}. & The architectures of piracy models have a limited impact on the performance of ME attacks, while more advanced models may not lead to more effective attacks. & $\LEFTcircle$ \\
    \hline
    Using more complex models tends to achieve better attack performance\mcite{shi2017steal,krishna2019thieves,aivodji2020model}. & The impact of model complexity varies with the concrete tasks (\mie, FER versus NLU) while pre-training is a more dominating factor.&$\LEFTcircle$ \\
    \hline
Semi-supervised learning substantially improves the query efficiency of ME attacks, especially when the query budget is low\mcite{jagielski2020high}.
& Semi-supervised learning improves the query efficiency of ME attacks, but the margin of improvement is not as large as reported in local experiments\mcite{jagielski2020high}; further, it has a negative impact on adversarial fidelity. &$\LEFTcircle$ \\
\hline
Active learning substantially improves the query efficiency of ME attacks\mcite{pal2020activethief}. & Active learning yields only marginal improvements in query efficiency and can, in some cases, have a negative effect.
& $\LEFTcircle$\\
\hline
Using adversarial examples improves both the query efficiency and the attack effectiveness\mcite{papernot2017practical, juuti2019prada,pengcheng2018query,yu2020cloudleak}. & Using adversarial examples has a limited or even negative impact on attack fidelity, but may improve adversarial fidelity. &$\Circle$ \\
\hline
Perturbing output confidence scores effectively mitigates ME attacks\mcite{orekondy2019prediction,lee2019defending}. & Output quantization weakens ME attacks but is not sufficiently effective. & $\LEFTcircle$ \\
\end{tabular}
\caption{Comparison of conclusions in prior work and \system ($\Circle$ -- inconsistent; $\LEFTcircle$ -- partially inconsistent).
\label{tbl:finding}}
\vspace{-6pt}
\end{table*}

\vspace{2pt}
-- Leveraging \system, we conduct an empirical study on leading MLaaS platforms (\mie, Amazon, Microsoft, \facepp, and Google) in the tasks of facial emotion recognition (FER) and natural language understanding (NLU). We show that today's real-world MLaaS APIs still exhibit significant vulnerability to ME attacks, while the characteristics of vulnerability vary greatly across different platforms and tasks.

\vspace{2pt}
-- We further examine the influential factors on the performance of ME attacks, including optimizers, training regimes, model architectures, and advanced attack strategies. Our evaluation leads to a set of interesting findings, many of which challenge the conclusions in prior work, as summarized in Table\mref{tbl:finding}. For instance, it is found that compared with other factors (\meg, optimizer), the piracy models have a limited impact on the attack performance, challenging the conventional notion that more advanced models lead to more effective attacks. Also, it is shown that adversarial examples offer little boost to or may even negatively impact the efficacy of ME attacks, which contradicts the findings in prior work. 

\vspace{2pt}
-- Finally, by integrating a longitudinal dataset containing over 1.7 million queries to leading MLaaS platforms spanning from 2020 to 2022, we conduct a retrospective study to characterize the vulnerability evolution of these MLaaS platforms. We discover some significant trends in vulnerabilities and identify the impact of model updates on ME attacks. Our analysis leads to some notable findings, including the influence of model updates on ME attack results and the potential lack of investment in model protections. Our contribution highlights the historical perspective of the evolving ME attack in MLaaS platforms, emphasizing the necessity for enhanced security measures and proactive defense against ME attacks.

We envision that the \system platform and our findings facilitate future research on ME attacks and shed light
on building MLaaS platforms in a more secure manner.

%% file: Fundamentals.tex
\vspace{-6pt}
\section{Background}

\subsection{Preliminaries}
We first introduce fundamental concepts and assumptions used throughout the paper.

%\vspace{2pt}
\vspace{2pt}
\textbf{Deep neural networks (DNNs)}\mcite{lecun2015deep} represent a class of ML models to learn high-level abstractions of complex data. In a predictive task, a DNN $f_\theta$ (parameterized by $\theta$) encodes a function $f_{\theta}: \gX \rightarrow \gY$, which maps an input $x \in \gX$ to a class $y \in \gY$. The training of $f_\theta$ often involves iteratively updating $\theta$ via algorithms such as stochastic gradient descent (SGD)\mcite{sgd}, aiming to minimize a loss function (\meg, cross-entropy) that measures the discrepancy between the model's prediction $f_\theta(x)$ and the ground-truth class $y$. Techniques such as data augmentation, batch normalization, and dropout may also be used during training to accelerate training or prevent overfitting. As their design may require significant engineering effects and their training may involve substantial data and compute resources, DNN models are often considered invaluable intellectual property in various contexts. 

% The increasing scale of DNNs leads to a rise in training costs and complexity, accentuating the worth of the learned weights $\theta$ or their resultant functionality \mcite{schmidhuber2015deep}. Hence, these weights, embodying significant computational investment, are frequently regarded as invaluable intellectual property.

\vspace{2pt}
\textbf{Knowledge distillation (KD)}\mcite{knowdiss,gou2021knowledge} is a process where a student model $f^\mathrm{student}$ is trained to mimic a teacher model $f^\mathrm{teacher}$. Typically, KD involves minimizing the discrepancy between the two models, which can be measured in responses (\mie, models' outputs)\mcite{knowdiss}, features (\mie, models' intermediate representations)\mcite{huang2017like,romero2014fitnets}, or relations (\mie, models' modeling of input relationships)\mcite{lee2018self,yim2017gift}. Formally, 
\begin{equation}
\min_\theta \E_{x \in \gD}\, \Delta( f_\theta^\mathrm{student}(x), f^\mathrm{teacher}(x))
\end{equation}
where $\Delta$ measures the discrepancy between two models with respect to a reference dataset $\gD$.

% KD involves three key components: 1) \textit{Knowledge} acquired by $f_\mathrm{student}$ in the forms of responses ($f_\mathrm{teacher}$'s outputs)\mcite{knowdiss}, feature ($f_\mathrm{teacher}$'s intermediate representations)\mcite{huang2017like,romero2014fitnets}, or relations ($f_\mathrm{teacher}$'s modeling of input relationships)\mcite{lee2018self,yim2017gift}; 2) \textit{Distillation Algorithm} used to transfer the knowledge from $f_\mathrm{teacher}$ to $f_\mathrm{student}$, typically by minimizing the discrepancy between the two models; 3) \textit{Teacher-Student Architecture:} The teacher model is usually a large, complex model that has been trained on a large amount of data, while the student model is a smaller, simpler model that is designed to be more computationally efficient.

% Knowledge distillation is applicable in various fields and can be extended to tasks like adversarial attacks\mcite{yu2020cloudleak,pal2020activethief,papernot2016distillation}, lightweight model \mcite{cho2019efficacy,boutros2022pocketnet,wang2023lightweight}, and data privacy and security \mcite{kim2022efficient,wu2022federated}.

\vspace{2pt}
\textbf{Model extraction (ME)} aims to infer a victim model's properties typically through black-box query access. The inferred information may include the model's architecture, (hyper)parameters, functionality, and other properties (\meg, attack vulnerability). The existing ME attacks can be roughly categorized as exact or approximate extraction.

Exact extraction aims to infer the victim model's properties {\em exactly}, for instance, architectures\mcite{oh2019towards}, hyperparameters\mcite{wang2019stealing}, and parameters\mcite{tramer2016stealing} (with respect to known architectures). For instance, equation-solving attacks\mcite{tramer2016stealing} allow the extraction of the exact parameters of (multi-class) logistic regression and multi-layer perceptron models. Approximate extraction aims to construct a piracy model similar to the victim model\mcite{yu2020cloudleak,pal2020activethief,tramer2016stealing,papernot2016distillation,gong2021inversenet}, which can be further divided based on its goal: 1) obtaining similar performance as the victim model (measured by accuracy) or 2) acquiring similar behavior as the victim model (measured by fidelity).

In the following study, we primarily focus on KD-based, approximate ME attacks, due to their general applicability and limited assumptions.

% Defenses against model extraction attacks fall into two categories: reactive and proactive. Reactive defenses\mcite{juuti2019prada}, like ownership verification and attack detection, identify attacks either in progress or after they've occurred. Ownership verification uses unique identifiers or watermarking, while attack detection monitors for signs of unauthorized access. Proactive defenses\mcite{zheng2019bdpl,lee2019defending,guan2022you,guan2022you,hitaj2018have,orekondy2019prediction}, on the other hand, aim to prevent attacks by altering the model's architecture, parameters, decision boundary, or performance. However, these defenses aren't foolproof and often work only under specific conditions, highlighting the need for further research to develop more robust defenses.

\subsection{Threat model}

We first define the threat model assumed in our study.
%\vspace{2pt}

\textbf{Adversary's objective --} The adversary's goal is to generate a piracy model $f_p$ that approximates the behavior and/or performance of the MLaaS backend model $f_v$. The agreement between $f_p$ and $f_v$ is measured by either accuracy or fidelity with respect to a testing dataset $\gD^\star$.

% \begin{equation}
% \label{tm:goal}
% \sE_{(x, y) \in  \gD}\, \mathbb{I}( \arg\max(f_v(x)) = \arg\max(f_p(x))),
% \end{equation}

%\vspace{2pt}
\textbf{Adversary's knowledge --} We assume a black-box setting, in which the adversary has little knowledge about the victim model $f_v$, including its architecture, (hyper)parameters, and the specific dataset used in its training. Yet, the adversary has access to a reference (unlabeled) dataset $\gD$.

%\vspace{2pt}
\textbf{Adversary's capability --} The adversary is able to access $f_v$ through the MLaaS API, which, for a given query input $x$, returns $f_v$'s prediction $f_v(x)$. Note that the queries are not restricted to real data and may include synthesized or adversarial inputs. Moreover, as MLaaS often charges users on a per-query basis, we assume the adversary has a limited query budget $n_\mathrm{query}$.

\begin{figure}[!t]
    \centering
    \includegraphics[width=0.46\textwidth]{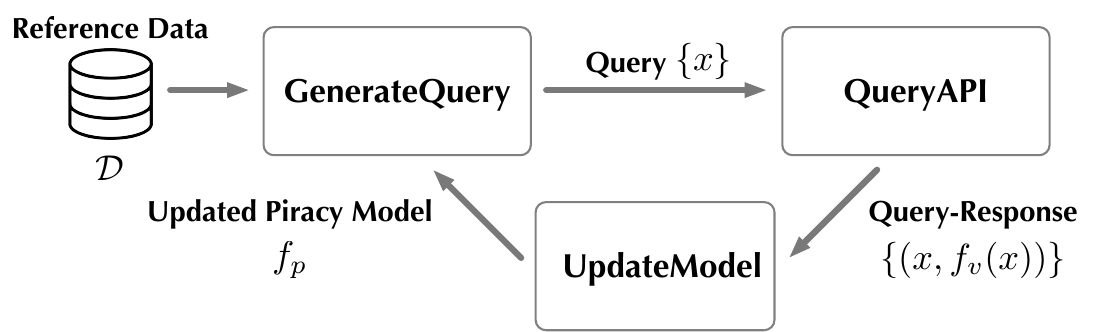}
    \caption{A general framework of ME attacks.}
    \label{fig:flow}
\end{figure}

\subsection{A general attack framework}
\label{sec:flow}

In \system, all the ME attacks are implemented within a general framework, as illustrated in Figure\mref{fig:flow}. This framework comprises three key functions i) {\sf GenerateQuery}, ii) {\sf QueryAPI}, and iii) {\sf UpdateModel}. Specifically, {\sf GenerateQuery} generates queries that are either sampled from the reference dataset $\gD$ or synthesized; {\sf QueryAPI} then sends each query $x$ to the MLaaS API and receives its prediction $f_v(x)$; finally, {\sf UpdateModel} optimizes the piracy model $f_p$ using the query-response pairs $\{(x, f_v(x))\}$. Notably, this process can be executed iteratively: {\sf GenerateQuery} generates queries based on both the updated piracy model and the previous query results (\mie, using active learning to identify the next batch of queries). 

In the default ME attack, we sample $n_\mathrm{query}$ inputs from $\gD$ to query the MLaaS API and use all the query-response pairs $\{(x, f_v(x))\}$ to train the piracy model $f_p$ by minimizing the following cross-entropy loss:
% We assume that the adversary has the dataset $D$, which may be derived from publicly available datasets or collected by the adversary itself, and the adversary has no way to know whether the contents of the dataset are used for training the model in MLaaS. The adversary selects some or all of the data in $D$, queries the MLaaS Api, and gets $(x,f_v(x))$ to generate a synthetic dataset $D_s$. Finally, the adversary uses the synthetic dataset $D_s$ to train the piracy model $f_p$ based on the optimization in Equation\mref{eq:crossentropy}. 
\begin{equation}
\label{eq:crossentropy}
  \min_\theta - \sum_{x\in \gD} f_v(x)^T \log{f_p(x;\theta)}
\end{equation}

%% file: Experiment.tex
\section{Status quo of ME vulnerability}

Leveraging \system, we conduct an empirical study on leading MLaaS platforms. Our study is designed to center around 4 questions: 

\noindent Q$_1$: How vulnerable are today's MLaaS APIs to ME attacks? 

\noindent Q$_2$: How do various factors influence such vulnerability? 

\noindent Q$_3$: How does the vulnerability evolve over time? 

\noindent Q$_4$: How effective are the defenses (if any) employed on the MLaaS platforms?

% These questions serve to investigate and elucidate the potential weaknesses and security concerns inherent in contemporary MLaaS APIs and their vulnerability to ME attacks.

\subsection{Experiment setting}

We begin by introducing the setting of our evaluation.

\vspace{2pt}
{\bf Datasets –} To be succinct yet reveal key issues, we primarily use 6 benchmark datasets: RAFDB\mcite{li2019reliable}, EXPW\mcite{SOCIALRELATION_2017},  KDEF\mcite{lundqvist2022karolinska}, FER+\mcite{BarsoumICMI2016}, IMDB\mcite{imdb}, and YELP\mcite{yelp}, spanning both the vision and natural language domains. %All of them are the popular datasets for the corresponding task on paper with code\mcite{paperwithcodefer,paperwithcodenlu}. 
The first 4 correspond to the facial emotion recognition (FER) task, which classifies given facial images into 7 possible expressions (\meg, ``calm''). The last 2 correspond to the natural language understanding (NLU) task, which classifies given text into 4 sentiments (\mie, ``positive'', ``negative'', ``neutral'', and ``mixed''). The details of the datasets are deferred to Table\mref{tbl:dataset}. 
By fault, following prior work, we partition each dataset into an 80\%/20\% split, designating the 80\% as the reference data $\gD$ (for ME attacks) and the remaining 20\% as the testing data $\gD^\star$ (for performance evaluation).
%assume the training set $\gD_\mathrm{trn}$ and testing set $\gD_\mathrm{evl}$ are disjoint subsets of the same dataset. We split the KDEF dataset into 80\% training set and 20\% test set, all other authors of the dataset have already split the training set and test set in advance and we follow their split. 
%By default, the size of training set $\gD_\mathrm{trn}$ of each dataset is set as follows: EXPW\,(27k), FER+\,(3K), KDEF\,(2.3K), RAFDB\,(12K), IMDB\,(25K), YELP\,(50K). 
In addition, we also evaluate the scenario in which  
$\gD$ and $\gD^\star$ come from different datasets in \msec{sec:trans}.

\vspace{2pt}
{\bf APIs --} 
We consider the APIs of 4 leading MLaaS service providers: Amazon, Google, Microsoft, and \facepp. By default, we assume the adversary has access to the complete query response (\mie, classification labels and confidence scores). Specifically, in the FER task, each query response contains the confidence scores of different expressions (\meg, ``calm''). In the NLU task, each query response includes the confidence scores of different sentiments (\meg, ``positive''). The details of each API are deferred to Table\mref{tbl:api}.

\vspace{2pt}
{\bf Piracy models –} In FER, we consider 4 representative architectures for the piracy model: VGG\mcite{simonyan2014very}, ResNet\mcite{he2015deep},  DenseNet\mcite{huang2017densely}, and ViT\mcite{vit}, while in NLU, we consider 2 Transformer-based architectures for the piracy model: RoBERTa\mcite{liu2019roberta} and XLNet\mcite{yang2019xlnet}. Using models of distinct architectures (\meg, residual blocks versus skip connects), we aim to factor out the influence of individual model characteristics. Besides the backbone model, we use one fully connected layer with softmax activation as the classification head. By default, we assume the piracy models are randomly initialized and trained from scratch. In \msec{sec:piracymodels}, we also consider the scenario in which the piracy models are pre-trained on public datasets. In the FER task, the models are pre-trained on the ImageNet-1K dataset; in the NLU task, the models are pre-trained on the BERT dataset\mcite{devlin2018bert}. 

\vspace{2pt}
{\bf Metrics –} To evaluate the effectiveness of ME attacks, we mainly use three metrics:

{\em Accuracy} measures the fraction of inputs in the testing set that are correctly classified by the piracy model $f_p$. Formally, 
\begin{equation}
\mathrm{Acc}(f_p) = \frac{\sum_{(x, y) \in \gD^\star}\mathds{1}_{f_p(x) = y}}{|\gD^\star|}
\end{equation}
where $\mathds{1}_A$ denotes the indicator function that returns 1 if the predicate $A$ is true and 0 otherwise.

{\em Fidelity} measures the fraction of inputs in the testing set that receive the same classification by the victim and piracy models\mcite{yu2020cloudleak,jagielski2020high,pal2019framework,correia2018copycat,papernot2017practical}. Formally,
\begin{equation}
\mathrm{Fid}(f_p) = \frac{\sum_{(x, y) \in \gD^\star}\mathds{1}_{f_v(x) = f_p(x)}}{|\gD^\star|}
\end{equation}

{\em Adversarial fidelity} measures the fraction of adversarial examples (with respect to $f_v$) that receive the same classification by $f_v$ and $f_p$. Formally, 
\begin{equation}
\textrm{AdvFid}(f_p) = \frac{\sum_{(x) \in \gA(\gD^\star)}\mathds{1}_{f_p(x) = f_v(x)}}{|\gA(\gD^\star)|}
\end{equation}
where $\gA$ is the adversarial attack (\meg, PGD\mcite{madry2018towards}) that generates adversarial examples with respect to $f_p$. We mainly use this metric in adversarial ME attacks (\msec{sec:advanceattack}).

\begin{table}[!ht]\small
\setlength\tabcolsep{5pt}
\renewcommand{\arraystretch}{1.2}
\centering
\begin{tabular}{c|cccc}
\multirow{2}{*}{API}   & \multicolumn{4}{c}{Dataset}\\
    \cline{2-5}
      &KDEF&RAFDB&EXPW&FER+\\
\hline
\hline
Amazon&+0.49$\pm$0.03&-0.98$\pm$0.15&-1.23$\pm$0.092&+1.99$\pm$0.12\\
Microsoft&/&+1.23$\pm$0.19&+2.12$\pm$0.17&+4.12$\pm$0.18\\
\facepp&+2.12$\pm$0.18&+1.78$\pm$0.13&+1.97$\pm$0.04&+0.45$\pm$0.14\\
\end{tabular}
\caption{Difference of attack fidelity w/ and w/o data augmentation. \label{tbl:dataaug}}
\vspace{-3pt}
\end{table}

{\bf Data augmentation --} In the vision domain, data augmentation (\meg, random cropping and resizing) plays a significant role in model training. Thus, we conduct a pilot study on its effectiveness in the basic ME attack, with results shown in Table\mref{tbl:dataaug}. Observe that data augmentation improves the attack fidelity in most cases. Thus, we apply data augmentation by default in the FER experiments.

\vspace{2pt}
{\bf ME attacks --} We evaluate a variety of ME attacks. The basic attack is based on knowledge distillation\mcite{yim2017gift}, which randomly samples $n_\mathrm{query}$ inputs from the reference dataset $\gD$ as queries, receives query response $f_v(x)$ from the MLaaS API for each query $x$, and subsequently trains the piracy model $f_p$ based on the query-response pairs $\{(x, f_v(x))\}$. Moreover, we take into account more advanced attacks that aim to minimize the number of queries through various strategies (\meg, semi-supervised learning) in \msec{sec:advanceattack}.

\subsection{Q1: Overall vulnerability} 
\label{sec:overall}
% \begin{table}[!ht]\footnotesize
% \setlength\tabcolsep{1.25pt}
% \renewcommand{\arraystretch}{1.5}
%     \centering
%     \caption{Attack vulnerability of different MLaaS FER and NLU APIs under the default setting.}
%     \label{tbl:opti}
%     \begin{tabular}{c|ccc|cc}
%     \multirow{2}{*}{API}&\multicolumn{5}{c}{Dataset}\\
% \cline{2-6}
% &RAFDB&FER+&EXPW&IMDB&Yelp\\
% \hline
% \hline
% Microsoft&84.24&79.78&84.38&86.63&92.18\\
% FacePP&66.76&62.28&68.39&/&/\\
% Google&68.58&61.8&70.1&/&/\\
% Amazon&64.09&63.2&71.09&83.69&90.88\\
%     \end{tabular}
% \end{table}

\begin{figure}[ht!]
    \centering
    \includegraphics[width=\linewidth]{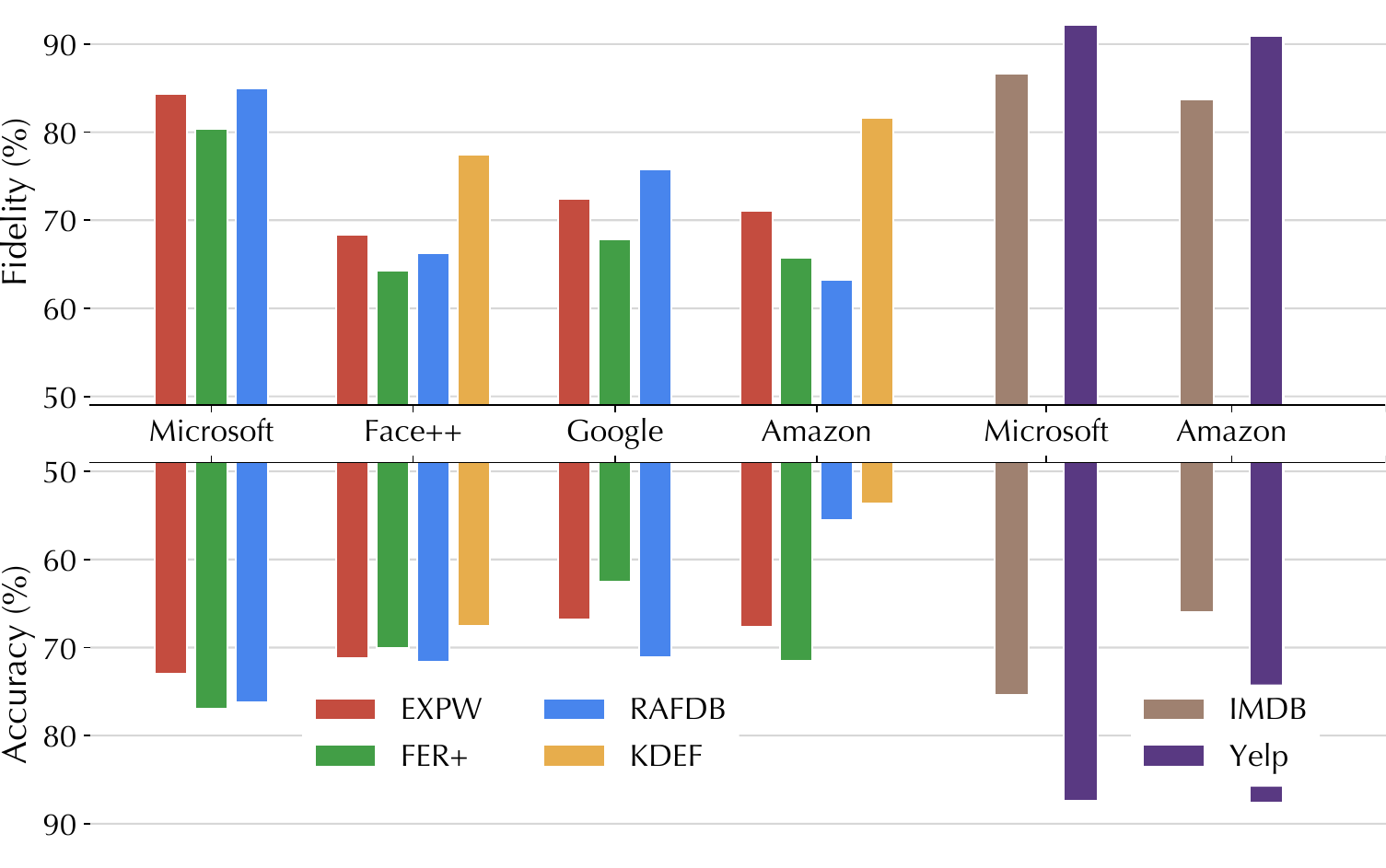}
    \caption{Vulnerability of different MLaaS APIs to ME attacks under the default setting.}
    \label{fig:exp1}
\end{figure}

% \begin{figure}[ht!]
%     \centering
%     \includegraphics[width=\linewidth]{fig/figure3_exp1.pdf}
%     \caption{Attack vulnerability of different MLaaS NLU APIs under the default setting. 
%     \rev{convert this figure to a table.}   
%     }
%     \label{fig:exp1_sa}
% \end{figure}

\subsubsection{Measurements}
We first examine the overall vulnerability of MLaaS APIs to ME attacks, with results summarized in Figure\mref{fig:exp1}. We have the following observations.
%not only assess the distinctive factors influencing such vulnerability but also examine the connections between ME evaluation metrics.

\vspace{2pt}
{\bf Platform variations --} A detailed evaluation of different APIs on the RAFDB dataset during ME attacks shows significant variations. The attack fidelity varies from 63.53\% against the Amazon API to  83.98\% against the Microsoft API. These variations can be attributed to several factors:

\vspace{1pt}
\mct{i} \textit{Label discrepancies} -- For instance, Amazon employs unique 8-class labels, diverging from the typical 7-class labels used by other platforms and the dataset itself.

\vspace{1pt}
\mct{ii} \textit{Training data distributions} -- While we lack precise knowledge about the original training data across different platforms, this factor is highly likely to influence the outcomes of ME attacks.

\vspace{1pt}
\mct{iii} \textit{Pre-processing and post-processing} -- The unique pre-processing (\meg, data cleaning, normalization, and augmentation) and post-processing (\meg, confidence score quantization) implemented by different MLaaS platforms may also have a substantial impact on the ME vulnerability.

\vspace{2pt}
{\bf Task/dataset variations --} Our findings also show that the ME vulnerability varies across tasks and datasets. This is evident in the significant fluctuation in the attack fidelity across different tasks and datasets. We attribute these variations to the unique characteristics intrinsic to each dataset.

In the NLU task, take the Amazon API as an example. Recall that the output of Amazon API falls into 4 categories (\mie, ``positive'', ``negative'', ``mixed'', and ``neutral''), where ``mixed'' exists as a separate class. IMDB is a high-polarity dataset, while Yelp is a diverse semantic dataset including a number of instances with mixed sentiments. This unique feature of the Yelp dataset allows the attack to better capture the ``mixed'' class, leading to higher attack fidelity compared with the IMDB dataset.

Compared with the NLU task, the FER task requires classifying given facial expressions into 7/8 emotion classes, without ambiguous ``mixed'' classes. This explains the relatively lower attack fidelity in the FER task. Further, observe that against the Amazon API, the attack fidelity on KDEF is much higher than the other datasets. This is explained by that KDEF explicitly instructs subjects to exhibit exaggerated expressions during data collection, which makes the classification task much easier. However, the piracy model extracted from KDEF does not generalize to other datasets (more details in \msec{sec:trans}). 

\vspace{2pt}
{\bf Fidelity vs. accuracy --} Figure\mref{fig:exp1} also shows that the variations in fidelity (63\% to 85\%) and accuracy (49\% to 77\%) across different APIs do not always align with each other. For instance, in the FER task, the attack against the Amazon API attains the highest fidelity yet the lowest accuracy on KDEF across different datasets. This is explained by that the agreement between the victim and piracy models, measured by fidelity, is not necessarily correlated with their performance, measured by accuracy. Given that ME attacks aim to extract the functionality of victim models, rather than attaining high performance, similar to prior work\mcite{gong2021inversenet,orekondy2019knockoff,pal2020activethief,yu2020cloudleak}, we primarily focus on the metric of attack fidelity in the study below.

%Thus, it is necessary to use multiple, complementary metrics to assess the effectiveness of ME attacks.

% High fidelity does not guarantee high accuracy. This phenomenon is all present in the FER task, which is closely related to the dataset characteristics of the FER mentioned in the previous paragraph, where a single ground-truth label is not a good and complete representation of the image's features. In addition, MLaaS uses a wide range of datasets for training, and its model outputs itself are not overfitted with a single dataset, and its distributional features are not consistent with a single dataset.

\vspace{2pt}
The findings above highlight the need for effective mitigation in mainstream MLaaS APIs. Moreover, it reiterates the importance of comprehending different factors (\meg, datasets and tasks) for accurately interpreting the ME vulnerability across different MLaaS platforms.

\subsubsection{Comparison with prior work}
We also compare our results with prior work on ME vulnerability in both settings of local models and MLaaS APIs.

\vspace{2pt}
{\bf Local models --} Most prior work (\meg,\mcite{gong2021inversenet,orekondy2019knockoff,pal2020activethief}) focuses on ME attacks against local models and uses CIFAR10 as the primary dataset. %InverseNet\mcite{gong2021inversenet}, KnockoffNet\mcite{orekondy2019knockoff}, ActiveThief\mcite{pal2020activethief}, and Papernot\mcite{papernot2016distillation} are the representative works. 
In particular, InverseNet\mcite{gong2021inversenet} shows superior performance over the other ME attacks on CIFAR10. If the adversary is able to access the full confidence information, InverseNet attains 45\%, 70\%, 81\%, and 82\% attack fidelity under 1K, 5K, 10K, and 15K queries, respectively. The fidelity growth plateaus after 15K queries, indicating a diminishing return from further increasing the query budget.

In Figure\mref{fig:exp1}, the Microsoft API shows the highest vulnerability, with 84\% attack fidelity for EXPW (27K queries) and RAFDB (12K queries), respectively. Due to its smaller size, the lower number of queries for FER+ (3K queries) results in slightly lower fidelity than the other two datasets. These findings corroborate with prior work, showing a positive positive correlation between query budget and attack fidelity.  

\vspace{2pt}
{\bf MLaaS APIs --} Among prior work, Cloudleak\mcite{yu2020cloudleak} conducts ME attacks against MLaaS APIs and measures attack fidelity with respect to different APIs and query budgets. Specifically, it reports 81\% fidelity and 58\% accuracy against the \facepp API on the KDEF dataset. Furthermore, it reports that the attacks against other platforms such as Microsoft and Clarifai, as well as different tasks including traffic sign recognition, flower, and NSFW, all attain over 82\% fidelity and 70\% accuracy. This clearly showcases the widespread nature of this vulnerability across various APIs and tasks. %\jiang{you claim that cloudleak does work in the following sections. here, you use it to show the vulnerability.} \jc{I think it's a problem}

In our experiments, on the same platform and dataset (\mie, KDEF and \facepp), we observe that the attack attains similar results (74\% fidelity and 66\% accuracy), while the performance gap with\mcite{yu2020cloudleak} may be attributed to the scratch model and the difference of testing data. In \mcite{yu2020cloudleak}, the testing data is synthesized, while we randomly sample 20\% of the KDEF dataset as the testing set. Further, similar to\mcite{yu2020cloudleak}, the ME vulnerability varies considerably across different APIs. In particular, the attack attains over 80\% fidelity against the Microsoft API across all the datasets, while the other APIs demonstrate more variations from one dataset to another.

Overall, we may conclude:
\begin{mtbox}{\small Observation\,1}{\small 
Many popular MLaaS APIs continue to be highly vulnerable to ME attacks, while this vulnerability varies greatly with concrete tasks and datasets.}
\end{mtbox}

% Since Microsoft has discontinued support for faceapi, the returned results from the Microsoft API provided by HAPI are used here. Note that only the confidence information of the highest confidence label is available in the HAPI dataset, for the other labels we will process the remaining confidence information according to the formula ? for the other tags, we will process the remaining confidence information according to Eq.

\subsection{Q2: Influential factors} 

Next, we evaluate how different key factors impact the effectiveness of ME attacks with respect to given MLaaS APIs.

% Lion(EvoLved Sign Momentum)\mcite{lion}, a novel optimization algorithm developed by Google Brain in 2023, is claimed to outperform Adam(w) by only tracking momentum and leveraging the sign function to calculate updates, resulting in reduced memory usage, accelerated model fitting and got a higher peak accuracy. We introduce the Lion optimizer to ME to compare with the traditional optimizer: AdamW, Adam and SGD.

\subsubsection{Optimizers}

Prior studies indicate that the selection of optimizers greatly impacts training dynamics\mcite{wu2021learning}. Thus, we examine how it affects ME attack performance.

We specifically consider 4 popular optimizers: SGD\mcite{sgd}, Adam\mcite{adam}, AdamW\mcite{adamw}, and Lion (EvoLved Sign Momentum)\mcite{lion}, which is a recently proposed optimizer. We configure the basic ME attack with varied optimizers and evaluate its effectiveness in the FER and NLU tasks.

\begin{table}[!ht]\small
\setlength\tabcolsep{2pt}
\renewcommand{\arraystretch}{1.2}
    \centering
    \begin{tabular}{cc|cccc}
     \multirow{2}{*}{API} & \multirow{2}{*}{Model}  & \multicolumn{4}{c}{Optimizer}\\
    \cline{3-6}
      &   & Lion & AdamW & Adam & SGD \\
        \hline
\hline
\multicolumn{6}{c}{RAFDB in FER}\\
\hline
Amazon&\multirow{2}{*}{ResNet50} &62.89$\pm$0.53& \cellcolor{Red}64.15$\pm$0.17& 63.96$\pm$0.06& 56.06$\pm$0.65\\
Microsoft&&\cellcolor{Red}84.64$\pm$0.20& 84.35$\pm$0.11& 84.35$\pm$0.10& 81.01$\pm$0.65\\
\hline

\multicolumn{6}{c}{IMDB in NLU}\\
\hline
\multirow{2}{*}{Amazon}&XLNet&\cellcolor{Red}84.30$\pm$0.16&82.45$\pm$0.20 &82.67$\pm$0.32&71.69$\pm$0.27  \\
&RoBERTa&\cellcolor{Red}81.20$\pm$0.21&74.72$\pm$0.08&74.58$\pm$0.12&42.93$\pm$0.32 \\
\hline

\multirow{2}{*}{Microsoft}&XLNet&86.35$\pm$0.23&\cellcolor{Red}86.57$\pm$0.52 &86.57$\pm$0.24&84.49$\pm$0.36\\
&RoBERTa&\cellcolor{Red}86.55$\pm$0.51 &85.20$\pm$0.54 &85.15$\pm$0.36 &65.75$\pm$0.29\\
    \end{tabular}
    \caption{Attack fidelity of different optimizers in FER and NLU. \label{tbl:optimizer}}
\vspace{-6pt}
    
\end{table}

\vspace{-6pt}
{\bf FER --} We compare different optimizers using RAFDB on the Amazon and Microsoft APIs.
Following\mcite{lion}, we set the hyper-parameters (\meg, learning rate $\mathrm{LR}$ and decay rate $\beta$) for different optimizers as follows: SGD with $\beta = (0.9,0.999)$ and $\mathrm{LR} = \mathrm{3e}{-2}$, Adam/AdamW with $\beta = (0.9,0.999)$ and $\mathrm{LR} = \mathrm{3e}{-4}$, and Lion with $\beta = (0.9,0.99)$ and $\mathrm{LR} = \mathrm{3e}{-4}$. We have the following key observations.

\begin{figure}[htbp]
\centering
\begin{subfigure}[b]{0.85\linewidth}
   \includegraphics[width=1\linewidth]{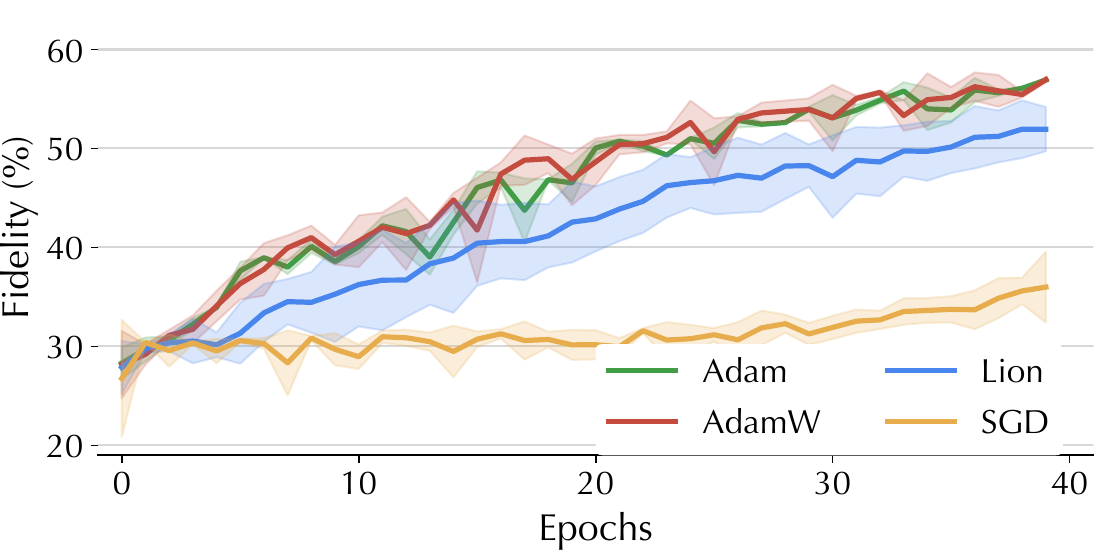}
   \caption{RAFDB on Amazon}

\end{subfigure}

\begin{subfigure}[b]{0.85\linewidth}
   \includegraphics[width=1\linewidth]{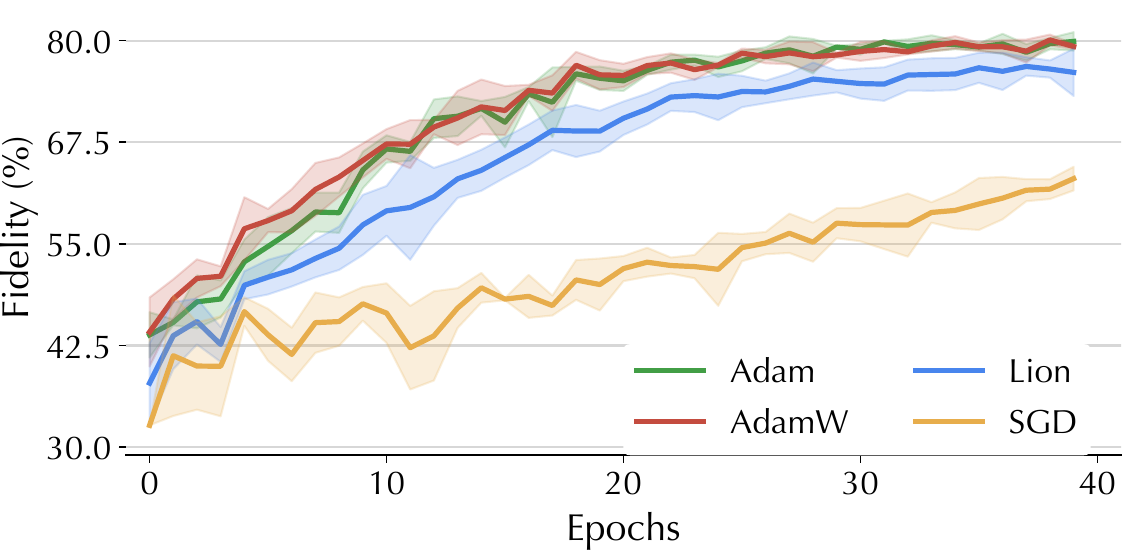}
   \caption{RAFDB on Microsoft}

\end{subfigure}

\caption{Training dynamics of different optimizers in the FER task.}
\label{fig:opti}
\end{figure}

Table\mref{tbl:optimizer} shows the attack fidelity at the end of 200 training epochs. Under the same query budget, Lion, Adam, and AdamW attain comparable attack fidelity. There is no distinct advantage between Lion and AdamW; however, it is evident that SGD lags behind the performance of other optimizers significantly.
%does not seem to be conducive to fitting in our scenarios, significantly lower than the other three optimizers.
We further examine the training dynamics of different optimizers. Figure\mref{fig:opti} summarizes how the attack fidelity varies with the number of training epochs. Despite the claimed superior convergence speed of Lion in supervised learning\mcite{lion}, we observe that Adam/AdamW actually converges much faster than Lion in the ME attack, while SGD converges fairly slowly.

% First, under the same query budget, all the optimizers eventually converge to similar attack fidelity, indicating that compared with other factors (\meg, query budgets), the optimizers may have a limited impact on the attack effectiveness in the FER task. 

% Second, Lion and AdamW consistently outperform the other optimizers in terms of convergence speed. Specifically, under KDEF+Amazon, Lion achieves a fidelity of $81.60\%$ around the 14-th epoch, which is substantially higher over SGD ($\mathrm{+}8.16\%$) and Adam ($\mathrm{+}9.20\%$), and converges in the first 25 epochs. Similarly, under RAFDB+Microsoft, AdamW is a slightly faster converge than Lion in the early stage, attaining the highest fidelity at the 37th epoch, outperforming both SGD ($\mathrm{+}1.56\%$) and Adam ($\mathrm{+}0.73\%$). Lion (84.43\% $\pm$0.53\%) and AdamW (84.29\% $\pm$0.65\%) eventually converge to similar attack fidelity.

\vspace{2pt}
{\bf NLU --} We also set the hyper-parameters of optimizers following\mcite{lion}: SGD with $\beta = (0.9,0.99)$ and $\mathrm{LR} = \mathrm{5e}{-4}$, Adam/AdamW with $\beta = (0.9,0.99)$ and $\mathrm{LR} = \mathrm{3e}{-6}$, and Lion with $\beta = (0.95,0.98)$ and $\mathrm{LR} = \mathrm{3e}{-6}$. 

Table\mref{tbl:optimizer} summarizes the attack performance corresponding to different optimizers on the IMDB dataset. Lion generally outperforms the other optimizers on both Microsoft and Amazon APIs. Our observations corroborate prior findings\mcite{lion} that Lion outperforms alternative optimizers in fine-tuning large language models. 

Overall, we may conclude:
% \vspace{2pt}
% Overall, although there are times when AdamW performs slightly better, Lion's consistent performance convinces us that he's the best first choice for ME attack optimizers.
\begin{mtbox}{\small Observation\,2}{\small
Advanced optimizers enhance ME attacks in terms of training efficiency and attack fidelity; the selection of optimizers depends on concrete datasets and tasks.}
\end{mtbox}

\vspace{2pt}
\subsubsection{Piracy models}
\label{sec:piracymodels}
The architectures of piracy models are also crucial for ME attacks. It is shown in prior work\mcite{chen2022teacher} that correctly identifying the architectures of victim models and using the same architectures in piracy models substantially boost the effectiveness of ME attacks. Several studies\mcite{orekondy2019knockoff,shi2017steal,krishna2019thieves,aivodji2020model} also demonstrate that the piracy model needs to be at least as complex as the victim model. Additionally, the results of \mcite{shi2017steal,krishna2019thieves,aivodji2020model} suggest that utilizing a more complex model leads to better attack performance. 

\begin{table}[!ht]\small
\setlength\tabcolsep{2pt}
\renewcommand{\arraystretch}{1.15}
    \centering
    \begin{tabular}{c|cccc}

Component& Model&Pre-training&\#Params\,(M)&Model Family\\
\hline
\hline

\multirow{5}{*}{\makecell{Model\\Family}}&GoogLeNet&\multirow{6}{*}{/}&6.7&GoogLeNet\\
&DenseNet&&8&DenseNet\\
&EfficientNet&&56&EfficientNet\\
&AlexNet&&61&AlexNet\\
&ResNet50&&25.6&ResNet\\

\hline

\multirow{4}{*}{\makecell{Model\\Complexity}}&ResNet18&\multirow{4}{*}{/}&11.7&\multirow{4}{*}{ResNet} \\
&ResNet50&&25.6&\\

&ResNet101&&44.6&\\
&WResNet50&&68.9& \\
\hline

\multirow{3}{*}{\makecell{Use of \\Pre-training}}&\multirow{3}{*}{ResNet50}&/&\multirow{3}{*}{25.6}&\multirow{3}{*}{ResNet}\\
&&ImageNet1K&&\\
&&VGGFace2&&\\
    \end{tabular}
 \caption{Setting of experiments on the impact of piracy models.}
    \label{tbl:piracymodel_detail}
\vspace{-8pt}
\end{table}

In the open-world setting targeted in this study, the victim models are not accessible, limiting us from directly comparing the architectures of piracy and victim models. Therefore, in the study below, we evaluate how different aspects of piracy models may impact the ME attacks with respect to given MLaaS APIs. Specifically, we explore three key aspects of model architectures: model families, model complexity (within the same family), and (non-)use of pre-training. Table\mref{tbl:piracymodel_detail} summarizes the experimental settings in evaluating the impact of piracy models.

\begin{figure}[!ht]
    \centering
    \includegraphics[width=0.8\linewidth]{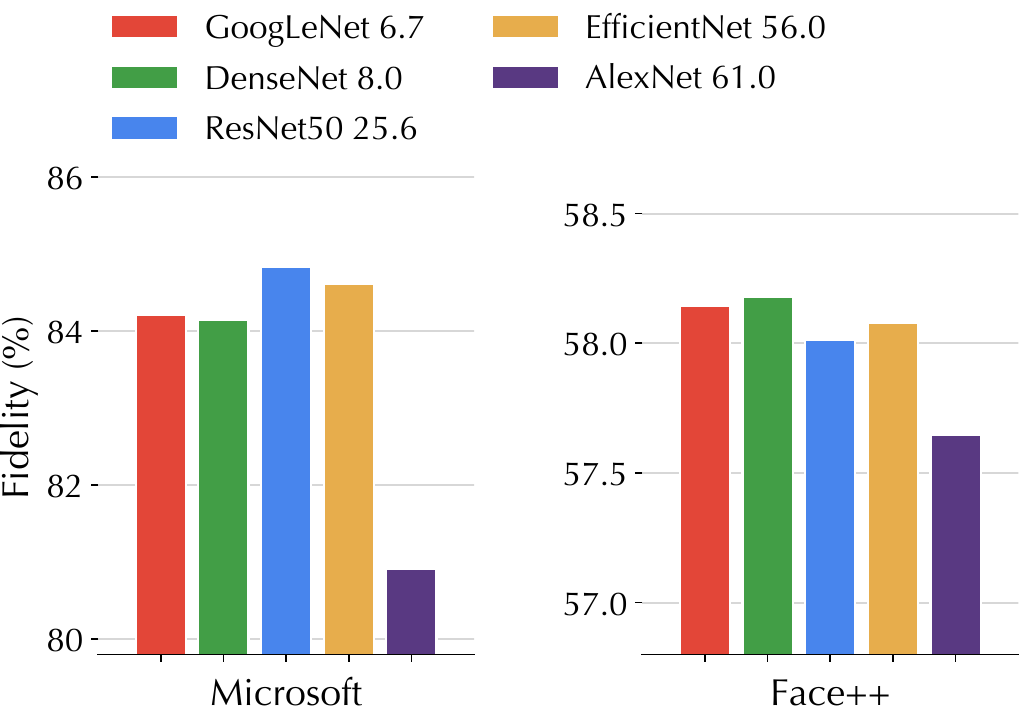}
    \caption{Impact of piracy model architectures on ME attacks (FER).}
    \label{fig:model_family}
\end{figure}

\vspace{2pt}
{\bf Model family --} For FER, we compare 5 popular architectures,  GoogLeNet\mcite{lenet}, DenseNet\mcite{huang2018densely}, ResNet\mcite{he2016deep}, EfficientNet\mcite{tan2020efficientnet}, Alex -Net\mcite{krizhevsky2014weird}, using RAFDB on the Microsoft and \facepp APIs, with results shown in Figure\mref{fig:model_family}. For NLU, we compare 2 popular LLMs, RoBERTa and XLNet, using IMDB on the Amazon and Microsoft APIs, with results summarized in Table\mref{tbl:surmodel-sa}. We have the following interesting findings.

\begin{table}[!t]\small
\renewcommand{\arraystretch}{1.15}
    \centering
    \begin{tabular}{cc|cc}
   \multirow{2}{*}{Model} & \multirow{2}{*}{Variant} & 
   \multicolumn{2}{c}{API} \\
   \cline{3-4}
& &    Amazon & Microsoft  \\
\hline
\hline
\multirow{2}{*}{XLNet} & Base & \cellcolor{Red}84.50  & 86.19 \\
& Large & 84.30 & \cellcolor{Red}86.35  \\
\hline
\multirow{2}{*}{RoBERTa} & Base & 78.96& 86.12 \\
& Large &\cellcolor{Red}81.20 & \cellcolor{Red}86.62\\
    \end{tabular}
     \caption{Impact of piracy model architectures on ME attacks (NLU).}
    \label{tbl:surmodel-sa}
\vspace{-8pt}
\end{table}

While prior work suggests that the piracy model architecture greatly impacts the effectiveness of ME attacks (\meg, 10-30\% difference in attack fidelity\mcite{chen2022teacher}). However, in our evaluation, except for AlexNet, an architecture proposed in 2012, which substantially under-performs the other architectures, the difference among other architectures is relatively marginal. A similar phenomenon is also observed in the NLU task as shown in Table\mref{tbl:surmodel-sa}.

It is worth pointing out that the number of parameters in these architectures varies from 6.7K to 61K. Yet, interestingly, GoogleLeNet, with only 6.7K parameters, delivers performance comparable with more complex architectures. To further evaluate the impact of model families, we consider ViT\mcite{vit}, a more advanced architecture with 86K parameters. However, it proves challenging to optimize ViT due to the limited data; even with pre-training, ViT does not match ResNet in terms of attack fidelity. This implies that for ME attacks, overly complex architectures can be counterproductive.

% \mct{ii} \jc{In addition, Figure \mref{fig:model_family} presents models with parameters ranging from 6.7K to 61K, but the performance difference between them is not significant, even GoogleLeNet (just with 6.7K parameters) can achieve good performance. In real experiments, we try with VGG (143.7K parameters) and ViT (86k parameters), and find that the models could not be optimized with a limited amount of data, and could not achieve the performance of ResNet even with the pre-trained models. This suggests that too many parameters and overly complex models are not conducive to ME attack.}    

\vspace{-3pt}
\begin{mtbox}{\small Observation\,3}{\small
In real-world MLaaS settings, the piracy model architecture has a limited impact on ME attacks, while more advanced architectures may not lead to more effective attacks.}
\end{mtbox}

% Our findings challenge the notion that more complex models (with a higher number of parameters) always perform better. For instance, the GoogleLeNet model, despite having the lowest parameter count, surpassed most other architectures on both APIs. Meanwhile, the widely-used ResNet architecture demonstrates consistent and moderate performance. It is noteworthy that\mcite{chen2022teacher} found that using the same model architecture as the teacher model led to a significant fidelity gain (10-30\%). However, in our experiments, the maximum difference in fidelity among different architectures was only 2.36\% and 1.40\%. This underscores that, in the context of extraction attacks, the attacker's choice of architecture may not significantly influence the results. As such, the classical ResNet architecture, known for its stable performance and minimal error rates, might serve as a reasonable and effective choice.

% \begin{table}[!ht]
% \setlength\tabcolsep{3pt}
%     \centering
%     \caption{Different model architecture in NLU}
%     \label{tbl:surmodel-sa}
%     \begin{tabular}{l|ccccc}
%         \toprule
%         \diagbox{Model}{Fidelity}{API} &amazon&microsoft&&&  \\
%         \midrule
%         XLNet-Base-Scratch& 36.3862(50.0801) & 65.7813(49.9199) &  & &  \\
%         XLNet-Base& 84.4992(84.8077) & 86.1939(76.1258) &  & &  \\
%         XLNet-Large & 84.2989(85.2925) &86.3542(77.8926) &  & &\\
%         Roberta-Base & 78.9583(84.9519) &86.1178(76.1899)  &  & &\\
%         Roberta-Large &81.1979(87.492)  & 86.6186(79.1827) &  & &\\

%         \bottomrule
%     \end{tabular}
% \end{table}

\begin{figure}[!ht]
    \centering        
\includegraphics[width=0.8\linewidth]{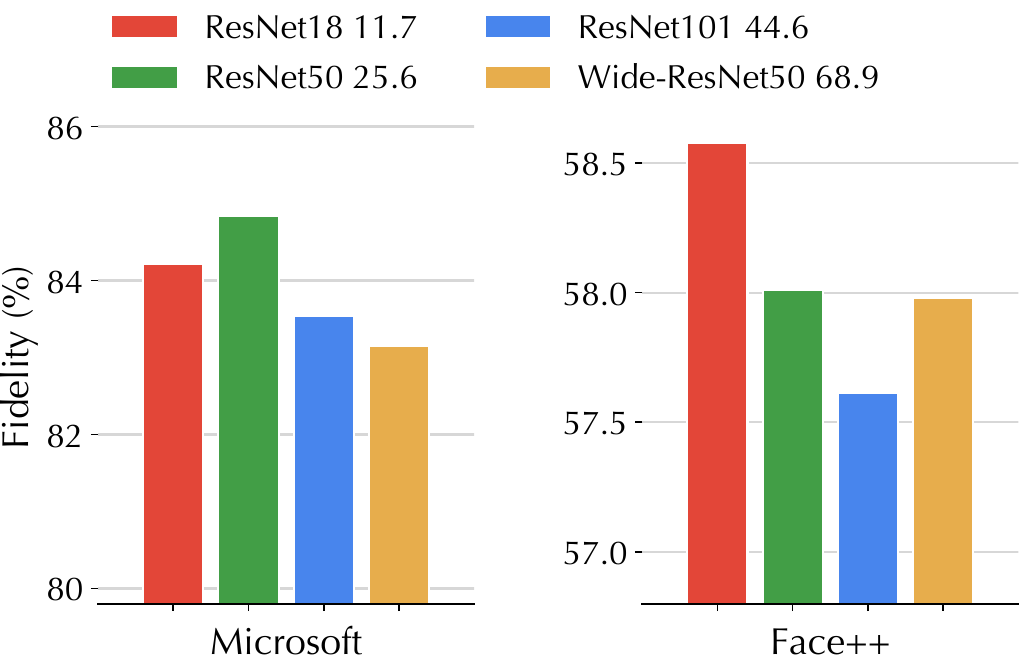}
        \caption{Impact of model complexity on ME attacks (FER).}
            \label{fig:model_complexity}
\end{figure}

{\bf Model complexity --} We further assess the impact of model complexity (within a single model family). In FER, we use ResNet as the piracy model and vary its width and depth, spanning across ResNet18, ResNet50, ResNet101, and Wide-ResNet50. As illustrated in Figure\mref{fig:model_complexity}, the attack fidelity across different architectures differs by less than $0.60\%$ and $0.20\%$ on Amazon and \facepp, respectively. Intuitively, as the model complexity reaches a certain level, further increasing the model width or depth does not lead to a notable improvement in attack performance. Similarly, in NLU, as shown in Table\mref{tbl:surmodel-sa}, for both RoBERTa and XLNet, using large models yields marginal improvement over base models. It is worth pointing out that compared with the base models, the large models of RoBERTa and XLNet are pre-trained using more data ($97\mathrm{GB}$ additional data)\mcite{yang2019xlnet,liu2019roberta}.

% Thus, the impact of model's width and depth on the attack performance is often intertwined with other factors such as pre-training data, which we explore next.
\begin{mtbox}{\small Observation\,4}{\small
Once reaching a certain level, further increasing the model complexity has a marginal impact on ME attacks.}
\end{mtbox}

%In fact, such changes were observed to potentially result in a slight decrease in fidelity. This suggests that tuning these parameters may not necessarily enhance the attack potential.

\vspace{2pt}
{\bf Use of pre-training} -- Next, we evaluate the influence of pre-training piracy models using public datasets on the performance of ME attacks.
% \jc{Considering that the pre-training set ImageNet used by the pre-trained model may contain corresponding datasets, we use the scratch Model in both FER tasks.  Since the dataset in the NLU task is small and does not allow for effective model fitting, we will use the pre-trained models and present their differences in Table\mref{tbl:surmodel-sa-pre}.}
% \jiang{Fidelity: pre-trained fune-tuning v.s. stealing?}

\begin{figure}[!ht]
    \centering
    \includegraphics[width=0.8\linewidth]{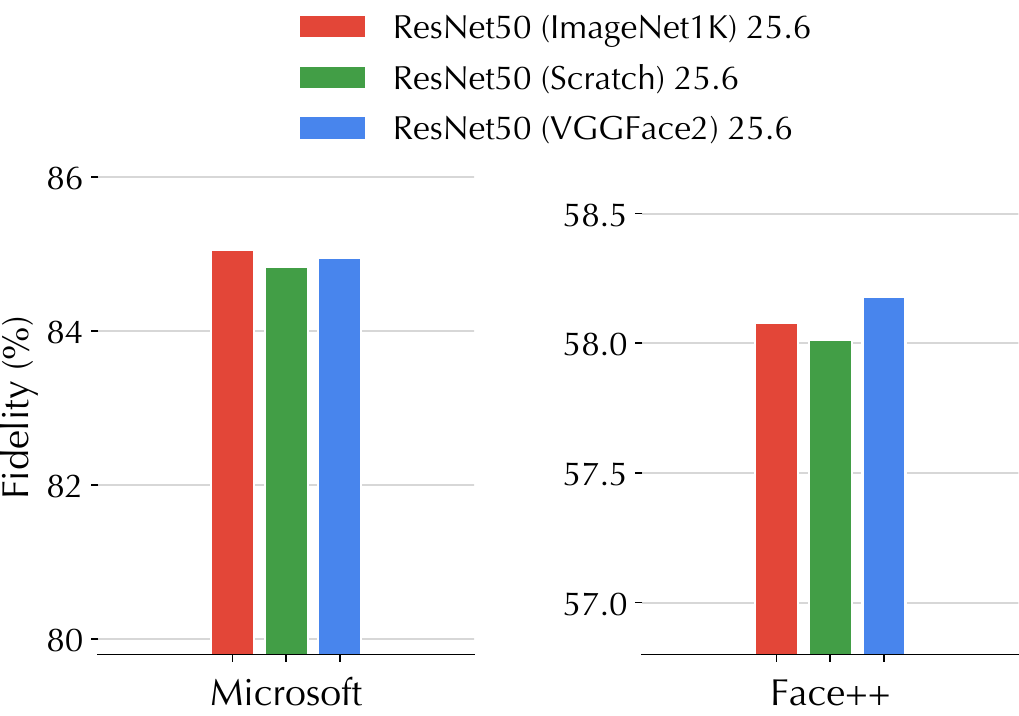}
        \caption{Impact of using pre-training on ME attacks (FER).}
        \label{fig:pre-training}
\end{figure}

\begin{table}[!ht]\small
\renewcommand{\arraystretch}{1.15}
    \centering
   
    \begin{tabular}{cc|cc}
   \multirow{2}{*}{Model} & \multirow{2}{*}{Pre-training} & 
   \multicolumn{2}{c}{API} \\
   \cline{3-4}
& &    Amazon & Microsoft  \\
\hline
\hline
 \multirow{2}{*}{XLNet-Base} & \ding{55} & 36.39 & 65.78\\
        & \ding{51} & \cellcolor{Red}84.50& \cellcolor{Red}86.19    \\
    \hline
        \multirow{2}{*}{RoBERTa-Base} & \ding{55} & 53.94& 73.61 \\
        &  \ding{51} & \cellcolor{Red}78.96  & \cellcolor{Red}86.11 \\
    \end{tabular}
     \caption{Impact of using pre-training on ME attacks (NLU). }
    \label{tbl:surmodel-sa-pre}
\vspace{-6pt}
\end{table}

In FER, we compare the attacks using a randomly initialized ResNet50 model as the piracy model with that using ResNet50 models pre-trained on public datasets. We consider two pre-training datasets, ImageNet-1K and VggFace2\mcite{cao2018vggface2}. As shown in Figure\mref{fig:pre-training}, pre-training marginally improves the attack performance, while the improvement margin varies with the APIs. For instance, compared with VGGFace2, ImageNet-1K leads to more improvement on Microsoft but less so on \facepp. In NLU, we compare randomly initialized XLNet and RoBERTa and that pre-trained on the BERT dataset\mcite{devlin2018bert}, with results shown in Table\mref{tbl:surmodel-sa-pre}. Surprisingly, unlike the PER task, pre-training language models significantly improves the attack performance. For instance, the pre-training of RoBERTa boosts fidelity by over 25\%. 
% \jc{Pre-training has limited improvement on basic and easy-to-optimize models like resnet, but on language models with a large number of parameters like the NLU task, pre-training becomes very important}
% \jiang{For simpler models like ResNet, pre-training offers minimal benefits. However, for large-parameter language models in the NLU task, pre-training is crucial.}

Thus, we may conclude:
\vspace{-3pt}
\begin{mtbox}{\small Observation\,5}{\small
Pre-training generally improves the performance of ME attacks, while the boost is more evident for language models than vision models.}
\end{mtbox}
\vspace{-3pt}

\begin{table*}[!ht]\small

\setlength\tabcolsep{3pt}
\renewcommand{\arraystretch}{1.2}
    \centering
     \begin{tabular}{ccc|c|ccccccc|c}
    \multirow{2}{*}{API}&\multirow{2}{*}{Dataset} & \multirow{2}{*}{Attack} &\multirow{2}{*}{Type} &
    \multicolumn{7}{c|}{Query Budget (\#\,Batches)}&\multirow{2}{*}{$ \overline\Delta$} \\
   \cline{5-11}
& && &   4 & 8 & 16 & 32 & 64 &128& Full &  \\

\hline
\hline

 \multirow{4}{*}{Microsoft}&\multirow{2}{*}{RAFDB} & Basic&\multirow{4}{*}{Fid} & 68.94$\pm$0.37	&73.8$\pm$0.88&	77.36$\pm$0.13	&79.43$\pm$0.33&	81.42$\pm$0.27	&83.29$\pm$0.16 &  \multirow{2}{*}{\makecell{84.35$\pm$0.13\\(191 Batches)}} & \multirow{2}{*}{+1.12} \\
 &&MixMatch &&\cellcolor{Red}70.25$\pm$1.35	&\cellcolor{Red}75.01$\pm$0.56&\cellcolor{Red}	78.67$\pm$0.16&	\cellcolor{Red}80.62$\pm$0.18	&\cellcolor{Red}82.47$\pm$0.35&\cellcolor{Red}	83.93$\pm$0.02 & \\

 &\multirow{2}{*}{EXPW} & Basic &&70.25$\pm$1.15&	73.42$\pm$0.42	&76.98$\pm$1.7&	78.45$\pm$0.72	&80.17$\pm$0.19	&81.54$\pm$0.31&\multirow{2}{*}{\makecell{83.94$\pm$0.08\\(417 Batches)}} & \multirow{2}{*}{+1.44}  \\
 &&MixMatch &&\cellcolor{Red}72.41(1.03)&	\cellcolor{Red}74.88$\pm$1.2&	\cellcolor{Red}77.65$\pm$1.33&	\cellcolor{Red}80.05$\pm$0.57&	\cellcolor{Red}81.6$\pm$0.03&	\cellcolor{Red}82.85$\pm$0.41& & \\

\hline

 \multirow{4}{*}{Amazon}&\multirow{4}{*}{FER+} & \multirow{2}{*}{Basic}&Fid &41.41$\pm$0.56	&48.09$\pm$1.22&	53.84$\pm$0.20	&59.66$\pm$0.32	&/&/&\multirow{4}{*}{\makecell{63.20$\pm$0.41\\18.57$\pm$1.12\\(50 Batches)}} & \multirow{4}{*}{+5.91}  \\

&&&AdvFid&\cellcolor{Blue}18.16$\pm$0.45&	\cellcolor{Blue}16.49$\pm$0.22	&\cellcolor{Blue}17.26$\pm$1.60	&\cellcolor{Blue}18.59$\pm$0.61&/&/& &\\

 && \multirow{2}{*}{MixMatch} &Fid&\cellcolor{Red}49.73$\pm$1.42&	\cellcolor{Red}55.39$\pm$1.40&\cellcolor{Red}	59.2$\pm$1.15&	\cellcolor{Red}62.32$\pm$0.16&/&/& & \\
 &&&AdvFid&  16.62$\pm$0.74&	16.05$\pm$1.20	&16.07$\pm$0.67&	15.7$\pm$1.07&/&/& &\\
    \end{tabular}
     \caption{Fidelity and adversarial fidelity of basic and MixMatch-based ME attacks (with batch size fixed as 64). In the case of MixMatch, each labeled batch is paired with an equal-sized unlabeled batch.}
  
\label{tbl:mixmatch}
\vspace{-6pt}
\end{table*}
\vspace{-3pt}

\subsubsection{Attack strategies}
\label{sec:advanceattack}

Thus far, we mainly focus on the basic attack strategy that randomly samples queries from the reference dataset. Next, we investigate advanced strategies that aim to minimize the number of queries. Specifically, we consider three popular strategies: semi-supervised learning, active learning, and adversarial learning. 

% \jiang{In previous sections, we delved into a basic attack strategy using random query inputs from reference data. Now, we'll investigate strategies that aim for fewer queries, focusing on semi-supervised, active, and adversarial learning.}

\vspace{2pt}
{\bf Semi-supervised learning --} Via semi-supervised learning, the adversary queries the API with a small number of samples and mixes the query responses (labeled data) and unlabeled data to train the piracy model. We consider MixMatch\mcite{mixmatch} as a representative semi-supervised learning method, which, in the context of ME attacks, is employed to reduce the number of queries\mcite{jagielski2020high}. 
At a high level, MixMatch-based ME attack samples a batch of samples to query the API as the labeled data $\gX$ and mixes it with an equally-sized batch of unlabeled data $\gU$ to produce a batch of augmented labeled data $\gX'$ and a batch of augmented unlabeled data with pseudo labels $\gU'$, which are used to train the piracy model but with different objective functions.

We evaluate the MixMatch-based ME attack in the FER task. Specifically, using the RAFDB and EXPW datasets on the Microsoft API, we compare the performance of basic and MixMatch-based ME attacks with varying numbers of queries, ranging from 4 to 128 batches (with batch size fixed as 64). Note that the FER+, RAFDB, and EXPW datasets include 50, 191, and 417 batches, respectively. As summarized in Table\mref{tbl:mixmatch}, we have the following interesting findings.

\vspace{1pt}
\mct{i} MixMatch generally enhances the effectiveness of ME attacks across different datasets and APIs. However, the improvement is less significant than that reported in\mcite{jagielski2020high}. For instance, it improves the fidelity by about 1\% against the Microsoft API and about 5\% against the Amazon API, under varying query budgets. The divergence from prior work may be attributed to both models and data. For instance, the results in\mcite{jagielski2020high} are evaluated using a one-layer fully-connected network on the CIFAR10 data, while the MLaaS backend models often use much more complex architectures trained on a massive amount of data (\meg, millions of images)\mcite{amazon-aws}. 

\vspace{1pt}
\mct{ii} Intriguingly, while boosting attack fidelity, MixMatch has a negative impact on adversarial fidelity, which measures the agreement between the victim and piracy models with respect to adversarial examples. For instance, the basic attack achieves adversarial fidelity 2.9\% higher than the MixMatch-based attack with $n_\mathrm{budget} =32$ batches. This observation may be explained as follows. MixMatch augments the training data by interpolating different samples, which essentially ``smooths'' the decision boundaries. While improving query efficiency, it also prevents the piracy model from effectively learning the high-curvature decision boundaries of the victim model, which is essential for adversarial fidelity.

\begin{mtbox}{\small Observation\,6}{\small
Semi-supervised learning generally improves attack fidelity under a limited query budget but has a negative impact on adversarial fidelity.}
\end{mtbox}

\vspace{2pt}
{\bf Active learning --} 
% To train the piracy model, instead of randomly querying the API, the adversary selects the most informative queries using active learning. This selection greatly reduces the need for data labeling, as the query responses are used to train the model.
The adversary may also employ active learning to choose informative inputs rather than randomly sampled ones to query the API to reduce the query cost.

We adopt $k$-Center-Greedy\mcite{sener2017active}, a method also used in Active -Thief\mcite{pal2020activethief}, in ME attacks. At each iteration, ActiveThief picks $k$ central samples from the unlabeled data (determined as cluster centers in the piracy model $f_p$'s latent space) to query the API. These responses are added to the training set $\gD$, which is used to train $f_p$. This process continues until the allocated query budget $n_\mathrm{budget}$ is exhausted. The results are summarized in Table\mref{tbl:active}. We have the following findings.
% We use $k$-Center-Greedy\mcite{sener2017active} as a representative active learning strategy, which, in the context of ME attacks, is employed in ActiveThief\mcite{pal2020activethief}. At each iteration, ActiveThief selects $k$ most representative samples (measured as $k$ 
% cluster centers in the latent space of the current piracy model $f_p$)
% from the unlabeled data to query the API inserts the query responses into the training set $\gD$, and then trains $f_p$ using $\gD$. The process iterates until the query budget $n_\mathrm{budget}$ is used up. The results are summarized in Table\mref{tbl:active}. We have the following findings.

\begin{table*}[!ht]\small
\setlength\tabcolsep{4pt}
\renewcommand{\arraystretch}{1.2}
    \centering
    \begin{tabular}{ccc|c|cccccc}

  \multirow{2}{*}{API} &  \multirow{2}{*}{Dataset} & \multirow{2}{*}{Attack} & \multirow{2}{*}{Type}&
    \multicolumn{6}{c}{Query Budget (\#\,Batches)} \\
   \cline{5-10}
& &   &&16&  32 & 64 & 96 & 128  & Full   \\
\hline
\hline

\multirow{4}{*}{Microsoft}&\multirow{2}{*}{RAFDB} & Basic&\multirow{4}{*}{Fid} & \cellcolor{Red}76.74$\pm$0.76 &	\cellcolor{Red}79.28$\pm$0.54	&81.76$\pm$0.08	&82.74$\pm$0.15	&83.2$\pm$0.39&  \multirow{2}{*}{\makecell{84.35$\pm$0.13\\(191 Batches)}} \\

&& ActiveThief  &&76.45$\pm$0.31&	79.17$\pm$0.4&	\cellcolor{Red}82.02$\pm$0.33 &\cellcolor{Red}82.93$\pm$0.3 &\cellcolor{Red}84$\pm$0.16 &  \\

&\multirow{2}{*}{EXPW} &Basic  &&\cellcolor{Red}75.94$\pm$0.56 &\cellcolor{Red}78.52$\pm$0.45 &\cellcolor{Red}79.76$\pm$0.08 &\cellcolor{Red}81.08$\pm$0.11 &\cellcolor{Red}81.57$\pm$0.24 &\multirow{2}{*}{\makecell{83.94$\pm$0.08\\(417 Batches)}}\\

&& ActiveThief & &74.56$\pm$0.68 &	77.09$\pm$0.42	&79.42$\pm$0.09	&80.43$\pm$0.47&	81.04$\pm$0.21&\\

\hline

\multirow{4}{*}{Amazon}&\multirow{4}{*}{RAFDB}&\multirow{2}{*}{Basic}&Fid&\cellcolor{Red}53.52$\pm$0.57 &\cellcolor{Red}57.07$\pm$0.61 &\cellcolor{Red}59.41$\pm$0.72 &\cellcolor{Red}61.08$\pm$0.8 &61.69$\pm$0.41 & \multirow{4}{*}{\makecell{63.53$\pm$0.14\\22.81$\pm$0.13\\(191 Batches)}}   \\

&&&AdvFid&\cellcolor{Blue}21.61$\pm$0.01 &\cellcolor{Blue}21.29$\pm$0.02 &\cellcolor{Blue}22.52$\pm$1.06 &	22.2$\pm$0.95 &\cellcolor{Blue}23.32$\pm$0.41 &\\

&& \multirow{2}{*}{ActiveThief}  &Fid& 52.14$\pm$0.07 &56.03$\pm$0.28 &	58.81$\pm$0.36 &	60.32$\pm$0.7 &\cellcolor{Red}61.81$\pm$1.17 &   \\
&&&AdvFid&20.34$\pm$0.6 &21.26$\pm$0.98 &	22.05$\pm$0.8 &	\cellcolor{Blue}22.99$\pm$0.95 &22.55$\pm$0.31 &\\

    \end{tabular}
    \caption{Fidelity and adversarial fidelity of basic and active learning-based ME attacks (with batch size fixed as 64).}
    \label{tbl:active}
    \vspace{-6pt}
\end{table*}

% \jc{When comparing the basic and active learning-based ME attacks on different APIs and datasets,  using active learning is even less effective than basic, both in terms of fidelity and attack fidelity data, especially when data queries are small.}

% \jiang{Compared to basic ME attacks, active learning-based attacks on various APIs and datasets are often less effective, showing reduced fidelity and attack fidelity, particularly with fewer data queries. }

% \jiang{this is counter-intuitive. any explanations?}

In our experiment combination of Microsoft+RAFDB, the results from Active Learning appear inconsistent. Some seed and query budgets show slight improvements, but these results are unstable, especially with lower query budgets. For instance, Active Learning performs worse with query budgets of 16 and 32 batches than the Basic attack. With a budget of 64 batches,  there is a slight average improvement of 0.26\%. However, as the query budget rises, the performance becomes steadier, although the improvement is never over 1\%. This initial inconsistency might be attributed to the initial model struggles to capture face feature maps well. When the model is capable, the number of queries has been exhausted.   As a result, when the model becomes more capable, the number of available queries has already been exhausted, which may prevent it from selecting the best data points. We notice a similar pattern with the Amazon+RAFDB combination. AdvFid and Fid perform worse than the Basic attack at low query budgets. Otherwise, in the Microsoft+EXPW combination, Active Learning always performs worse than the Basic attack.

Back to previous research on active learning, when $k$-Center-Greedy combines with data augmentation \mcite{mittal2019parting}, results in only slight accuracy improvements. Specifically, they show an increase of approximately 1 to 2\% on CIFAR10 compared to random sampling-based queries. In earlier studies related to ME attacks, \mcite{pal2020activethief} demonstrates that using the $k$-Center-Greedy method leads to an average improvement of 0.976\% in attack fidelity on CIFAR10. However, when the query budget increases to 25K, this method under-performs by -0.62\%. In previous work, it is effective in most cases but has limited improvement and shows instability.

In conclusion, active learning-based attacks on different APIs and datasets often perform worse than basic ME attacks. They demonstrate lower fidelity and adversarial fidelity, especially when there are fewer data queries. Given its inconsistent performance, active learning is not recommended for ME attacks in real-world scenarios.

% \jiang{please explain your results. Currently, all the discussion focuses on the previous work.}

% \jc{it's hard to make an explanation here}
% \citet{mittal2019parting} studies the effect of active learning on a typical CIFAR10 classification task. They observe that when active learning algorithms, including k-Center-Greedy, were applied along with data augmentation, the improvement yielded by active learning is quite limited. In particular, when the data amount is between 5k and 20k, incorporating active learning algorithms resulted in approximately a 1\% to 2\% performance improvement, suggesting that the weak role of using active learning algorithms. 

% \mct{ii} Activethief \mcite{pal2020activethief} incorporated active learning into ME, but their work did not include real API experiments. Their results showed that when the k-Center-Greedy algorithm was utilized, the average boost under CIFAR10 was 0.976\%. Nevertheless, when the data size reached 25k (21\%), the performance of the k-Center-Greedy algorithm declined and underperformed random selection by -0.62\%. The same phenomenon is presented in our mlaas experiment: The effect of active learning is weak and inconsistent. In some instances, active learning underperformed compared to random data selection. In summary, it might not be necessary to expend significant resources to run active learning algorithms given their limited effectiveness.

\begin{mtbox}{\small Observation\,7}{\small
Introducing active learning is likely to acquire negative effects, especially when the amount of queries is small}
\end{mtbox}

\label{sec:AdversarialExample}
{\bf Adversarial learning --} 
CloudLeak\mcite{yu2020cloudleak} applies adversarial learning to generate examples to reduce the required number of queries in ME attacks. Specifically, the strategy is to perform adversarial attacks on clean data $\gX$ and produce adversarial examples $\gA$ on the piracy model $f_p$. These examples represent highly uncertain regions of $f_p$. When these regions are queried against the API, they offer valuable insights to refine $f_p$. Additionally, since $\gA$ lies close to $f_p$'s decision boundaries, these queries push $f_p$ towards the decision boundaries of the victim model $f_v$.
We consider two representative adversarial attacks: PGD\mcite{madry2018towards} (with $\epsilon = 4/255 $, $ \alpha = 2/{255} $, $ n_\mathrm{iter} = 7 $, and random $ \delta $ initialization) and CW (with $\kappa = 40 $, $ n_\mathrm{step} = 50 $, and $ \mathrm{LR} = 0.01 $)\mcite{carlini2017towards}.  The results are summarized in Table\mref{tbl:adv}.

\begin{table}[!ht]\small
\setlength\tabcolsep{3pt}
\renewcommand{\arraystretch}{1.1}

    \centering
    \begin{tabular}{c|c|c|ccc}
      \multirow{2}{*}{API}& \multirow{2}{*}{Dataset}& \multirow{2}{*}{\makecell{Query\\Batches}} & 
    \multicolumn{3}{c}{Attack} \\
       \cline{4-6}
  &&&   Basic & PGD & CW      \\
\hline
\hline
\multirow{12}{*}{Amazon}&\multirow{6}{*}{RAFDB}&\multirow{2}{*}{8}&\cellcolor{Red}40.46$\pm$0.39&36.97$\pm$0.1&37.5$\pm$0.22\\
&&&17.27$\pm$0.27&19.02$\pm$0.05&\cellcolor{Blue}20.35$\pm$0.35\\

&&\multirow{2}{*}{16}&\cellcolor{Red}43.48$\pm$0.47&41.66$\pm$0.17&40.39$\pm$0.45\\
&&&19.66$\pm$0.17&19.16$\pm$0.29&\cellcolor{Blue}21.27$\pm$0.15\\
&&\multirow{2}{*}{32}&45.94$\pm$3.81&42.92$\pm$2.02&\cellcolor{Red}46.91$\pm$2.94\\
&&&21.33$\pm$0.15&20.71$\pm$0.22&\cellcolor{Blue}23.4$\pm$0.48\\
\cline{2-6}
&\multirow{6}{*}{FER+}&\multirow{2}{*}{8}&35.32$\pm$1.83&39.91$\pm$0.2&\cellcolor{Red}41.52$\pm$1.58\\
&&&17.06$\pm$0.01&\cellcolor{Blue}18.86$\pm$0.97&18.31$\pm$1.28\\
&&\multirow{2}{*}{16}&41.5$\pm$1.83&42.33$\pm$0.02&\cellcolor{Red}46.4$\pm$3.97\\
&&&19.02$\pm$0&\cellcolor{Blue}20.81$\pm$0.08&17.75$\pm$4.91\\
&&\multirow{2}{*}{32}&44.56$\pm$0.2&49.35$\pm$0.38&\cellcolor{Red}50.29$\pm$0.32\\
&&&17.87$\pm$0.74&\cellcolor{Blue}19.46$\pm$1.34&17.28$\pm$0.35\\
\hline

\multirow{12}{*}{\facepp}&\multirow{6}{*}{RAFDB}&\multirow{2}{*}{8}&\cellcolor{Red}40.92$\pm$0.41&40.92$\pm$1.27&35.07$\pm$2.7\\
&&&20.31$\pm$0.29&\cellcolor{Blue}20.79$\pm$0.09&20.56$\pm$0.14\\
&&\multirow{2}{*}{16}&\cellcolor{Red}44.65$\pm$3.25&40.01$\pm$0.19&41.09$\pm$0.07\\
&&&\cellcolor{Blue}18.58$\pm$0.34&16.57$\pm$0.48&17.15$\pm$0.05\\
&&\multirow{2}{*}{32}&\cellcolor{Red}47.91$\pm$0.46&45.28$\pm$0.18&45.48$\pm$0.29\\
&&&15.54$\pm$0.21&16.2$\pm$0.43&\cellcolor{Blue}17.29$\pm$0.24\\
\cline{2-6}
&\multirow{6}{*}{FER+}&\multirow{2}{*}{8}&\cellcolor{Red}67.79$\pm$0.31&22.58$\pm$1.8&51.02$\pm$0.01\\
&&&14.46$\pm$0.2&\cellcolor{Blue}23.98$\pm$0.46&18.74$\pm$0.3\\
&&\multirow{2}{*}{16}&\cellcolor{Red}72.61$\pm$3.39&45.6$\pm$0.49&41.65$\pm$2.02\\
&&&14.46$\pm$0.46&\cellcolor{Blue}18.28$\pm$0.2&13.2$\pm$0.2\\
&&\multirow{2}{*}{32}&\cellcolor{Red}67.79$\pm$0.13&47.51$\pm$0.01&47.51$\pm$0.27\\
&&&14.46$\pm$0.47&\cellcolor{Blue}17.79$\pm$0.01&17.57$\pm$0.22\\

    \end{tabular}
    \caption{Fidelity and adversarial fidelity of basic and adversarial learning-based ME attacks (with batch size fixed as 64).}
     \label{tbl:adv}
\end{table}

% However, it is observed that under the same query budget, compared with querying the API with clean data, using adversarial learning to generate queries does not improve the performance of ME attacks consistently. On the contrary, in most cases, using clean data to query the API leads to higher attack fidelity. 

% \jiang{need re-organize}
% Cloudleak \mcite{yu2020cloudleak} uses \facepp and KDEF dataset as the experiment. KDEF is a small amount dataset and includes the exaggerated facial expressions made by subjects. They report attack fidelity of $85.99\%$ (PGD) and $98.74\%$ (CW) with 2K queries (equivalent to 32 batches), compared to $80.58\%$ (basic) using clean data. We try to reproduce this experiment to see the effect of attack fidelity and adversarial fidelity, but CloudLeak uses a self-created testing set to evaluate the ME attacks. We were not able to obtain the testing set.

It is observed that against the Amazon API, using PGD or CW to enhance ME attacks leads to a considerable improvement in attack fidelity (around 5.61\% on average) on FER+, while the effect on RAFDB is fairly mixed. Meanwhile, against \facepp, adversarial learning-enhanced ME attack seems not only ineffective but even under-performs the basic attack. However, across all the cases, adversarial learning leads to a consistent improvement in adversarial fidelity. Our findings indicate that clean queries (sampled from the reference dataset) and adversarial queries (generated by adversarial attacks) seem to contribute to ME attacks differently, respectively improving the fidelity and adversarial fidelity of piracy models.

\begin{table}[!ht]\small
\setlength\tabcolsep{5pt}
\renewcommand{\arraystretch}{1.2}
    \centering
   
% \begin{tabular}{c|m{1cm}<{\centering}|m{1cm}<{\centering}|m{1cm}<{\centering}|m{1cm}<{\centering}}
    \begin{tabular}{c|c|ccccc}

       \multirow{2}{*}{Dataset}  &\multirow{2}{*}{Type} &\multicolumn{5}{c}{Adversarial/Clean Ratio $|\tilde{\gX}| :|\gX|$}\\
           \cline{3-7}
        && 1:0 &3:1&1:1&1:3& 0:1 \\
        \hline
        \hline

\multirow{2}{*}{RAFDB}&Fid&42.92&44.85&43.78&45.41&\cellcolor{Red}45.94\\
&AdvFid&20.71&20.57&20.55&20.37&\cellcolor{Blue}21.33\\
\hline
\multirow{2}{*}{FER+}&Fid&\cellcolor{Red}49.35&44.2&43.14&47.00&44.71\\
&AdvFid&19.46&18.62&20.06&\cellcolor{Blue}20.07&17.35\\

    \end{tabular}
     \caption{Fidelity and adversarial fidelity under varying adversarial/clean ratios in queries.}
 \label{tbl:adv_pgd}
 \vspace{-3pt}
\end{table}

These observations lead us to explore an interesting question: whether combining clean and adversarial inputs in querying the MLaaS API improve the overall efficacy of ME attacks. To investigate this, we vary the ratio of clean $\gX$ and adversarial $\gA$ (generated by PGD) queries in each batch while keeping the query budget constant. The experimental results are summarized in Table\mref{tbl:adv_pgd}.

In some cases (\meg, RAFDB), including adversarial queries, may negatively impact the attack fidelity. In other cases (\meg, FER+), although only using adversarial queries leads to the best fidelity, the proportion of adversarial queries is not consistently correlated with the attack fidelity, which conflicts with the findings in\mcite{yu2020cloudleak}. Similarly, although using more adversarial queries improves adversarial fidelity,   increasing the proportion of adversarial queries is not strictly correlated with the increase of adversarial fidelity.

It is possible to explain these observations as follows. There exists intricate dynamics between clean and adversarial queries: while adversarial queries often represent corner cases with respect to the model's decision boundaries, clean queries represent more average cases. Thus, while increasing the number of queries generally improves the agreement between the victim and piracy models, clean and adversarial queries have different focuses.

%The alternative explanation is that there may be a denoising mechanism in the APIs invalidating our adversarial queries; however, we cannot verify this.

The results highlight the complexity of operating this attack strategy as well as its dependency on concrete datasets and APIs. Based on the findings, further research is needed to comprehend the possible advantages of applying adversarial learning in ME attacks and the optimal strategy to combine clean and adversarial queries to achieve high attack efficacy.

%Based on the findings, we are of the opinion that further research is required to completely comprehend the possible advantages of applying adversarial learning in ME attacks. 
%Our results highlight the complexities of these attacks and the challenges in replicating results across various APIs and dataset. 

% We make some modifications to the pseudocode provided by CloudLeak to attempt to improve our results. In CloudLeak, the synthetic dataset is entirely formed from adversarial samples. However, our results indicate that adversarial samples may bring side effects to the task, prompting us to attempt to replace some of the adversarial samples with clean samples. This experiment uses Projected Gradient Descent (PGD) to generate adversarial samples.

% we incrementally increase the proportion of clean samples $x$ to observe changes in fidelity. For the KDEF dataset, fidelity first decreases and then continuously increases as the proportion of clean samples rises, until it approaches the results obtained when using only clean samples. For RAFDB, it peaks when the proportion is 1:1. However, this result is very close to that obtained when using only clean samples (\mie, only 0.23\% lower), and it is 2.49\% higher than when using only adversarial samples $s$. This clearly demonstrates that, in this setting, adversarial samples do not further enhance the quality of the model, and indeed, they are outperformed by clean samples.

Overall, we may conclude:
\begin{mtbox}{\small Observation\,8}{\small
It is essential to properly adjust the proportions of clean and adversarial queries in ME attacks to optimize different metrics.}
\end{mtbox}

\subsection{Q3: Generalizability}
%\jiang{Reference data D, evaluation dataset $D^\star$}

\label{sec:trans}
Generalizability is a crucial metric for ME attacks: it measures how the extracted piracy model agrees with the victim model on datasets other than that used in the attack. In other words, generalizability indicates how valuable and flexible the extracted piracy model is. Next, under the MLaaS setting, we empirically measure the generalizability of piracy models. We extract the piracy model $f_p$ by querying the API using the reference dataset $\gD$ and evaluate $f_p$'s fidelity and accuracy on another dataset $\gD^\star$.

\subsubsection{FER} 
We conduct experiments across four datasets on three APIs to analyze the generalizability of ME attacks in the FER task, with results summarized in Table\mref{tbl:trans_fer}. The highlighted cells indicate when the original and evaluation datasets are identical, serving as a baseline. The results under Microsoft+KDEF are unavailable due to the closure of the Microsoft API. We have the following key observations from Table\mref{tbl:trans_fer}.

\begin{table}[htbp]\small
\setlength\tabcolsep{4pt}
\renewcommand{\arraystretch}{1.1}
    \centering
    
    \begin{tabular}{c|c|cccc}

          \multirow{2}{*}{Evaluation Dataset} &  \multirow{2}{*}{API} & 
    \multicolumn{4}{c}{Origin Dataset} \\
   \cline{3-6}
& &   KDEF&   RAFDB & EXPW & FER+       \\

    \hline
    \hline
\multirow{3}{*}{KDEF}&Amazon&\cellcolor{Red}77.78&59.90&67.36&55.21 \\
&Microsoft&\cellcolor{Red}/&/&/&/ \\
&\facepp&\cellcolor{Red}73.61&63.72&67.01&50.00 \\
\hline
\multirow{3}{*}{RAFDB}&Amazon&33.64&\cellcolor{Red}63.53&57.38&48.34 \\
&Microsoft&/&\cellcolor{Red}85.01&83.98&75.17 \\
&\facepp&37.67&\cellcolor{Red}66.76&66.02&57.15 \\
\hline
\multirow{3}{*}{EXPW}&Amazon&37.35&48.03&\cellcolor{Red}71.09&53.14 \\
&Microsoft&/&72.75&\cellcolor{Red}84.31&77.09 \\
&\facepp&47.79&56.91&\cellcolor{Red}68.40&58.91 \\
\hline
\multirow{3}{*}{FER+}&Amazon&33.42&49.68&65.85&\cellcolor{Red}63.2 \\
&Microsoft&/&72.80&83.13&\cellcolor{Red}80.36 \\
&\facepp&40.78&56.06&66.39&\cellcolor{Red}64.29 \\
    \end{tabular}
    \caption{Generalizability of different datasets and APIs (FER).}
    \label{tbl:trans_fer}
    \vspace{-6pt}

\end{table}

\vspace{2pt}
{\bf Asymmetric generalizability --} A piracy model extracted from dataset $A$ may perform well on dataset $B$, but the reverse is not always true. For instance, the model extracted from the EXPW dataset demonstrates strong generalizability across datasets, especially matching or surpassing baselines on RAFDB and FER+. However, the models extracted from RAFDB and FER+ do not perform as well when applied to EXPW. We speculate that highly generalizable datasets, such as EXPW, often benefit from their larger query volumes, allowing them to capture the distributions of smaller, less adaptable datasets like RAFDB and FER+. Specifically, with its 27K queries, EXPW substantially overshadows RAFDB's 12K and FER+'s 3K. Given their shared tasks, EXPW's comprehensive nature likely contributes to its superior generalizability.

{\bf Generalizability beyond query budgets --} Though KDEF and FER+ boast similar query volumes, KDEF's generalizability lags significantly. This can be attributed to its unique data collection methodology, where subjects were instructed to exhibit exaggerated emotions. This induces a distribution shift in KDEF, making it deviate from the normative distributions of facial image datasets. Conversely, RAFDB, FER+, and EXPW, derived from normal photos, have distributions that are more congruent with one another.

\subsubsection{NLU} 

Table\mref{tbl:trans_sa} shows the generalizability results in the NLU task. ME attacks using IMDB as the reference dataset $\gD$  demonstrate strong generalizability, particularly on Microsoft, where fidelity reaches 90.49\% compared to the baseline of 92.18\% when transferred to Yelp. However, the generalizability is weaker on Amazon, with a fidelity of 80.71\% compared to the baseline of 90.88\%. The results of Yelp are interesting, showing poor generalizability to Amazon (similar to IMDB) at 68.05\%, but high generalizability to Microsoft like IMDB.

In general, the generalizability of ME attacks in NLU is related to the APIs, Amazon is weaker than Microsoft, possibly because Amazon uses 4 classes compared to Microsoft uses 3 classes. Further, IMDB's data is highly polarized, and Yelp's data also contains neutral and mixed sentiments, which might also contribute to the results of generalizability.

% \begin{table*}[!ht]
% \renewcommand{\arraystretch}{1.2}
%     \label{tbl:adv}
%     \centering
%     \caption{Comparison of performance on victim models and their local substitute models}
%     \begin{tabular}{c|c|c|ccc|c}

\begin{table}[!ht]\small
\renewcommand{\arraystretch}{1.15}
    \centering

    \begin{tabular}{c|c|cc}

          \multirow{2}{*}{Evaluation Dataset} &  \multirow{2}{*}{API} & 
    \multicolumn{2}{c}{Origin Dataset} \\
   \cline{3-4}
& &   IMDB&   Yelp       \\

\hline
\hline

\multirow{2}{*}{IMDB}&Amazon&\cellcolor{Red}84.30&68.05 \\
&Microsoft&\cellcolor{Red}86.35& 84.60\\
\hline
\multirow{2}{*}{Yelp}&Amazon&  80.71&\cellcolor{Red}90.88 \\
&Microsoft& 90.49&\cellcolor{Red}92.18 \\
    \end{tabular}
     \caption{Generalizability of different datasets and APIs of NLU}
    \label{tbl:trans_sa}
    \vspace{-6pt}
\end{table}

    \vspace{-6pt}

\begin{mtbox}{\small Observation\,9}{\small
The generalizability of piracy models depends on the relationships between original and evaluation datasets, much more than other factors (APIs, victim models, ME attacks).}
\end{mtbox}
    \vspace{-6pt}

\subsection{Q4: Defenses}
\label{sec:quanti}
It is unclear what specific defense mechanisms are employed by the MLaaS platforms. However, when examining their APIs, Amazon, Microsoft, and \facepp will provide exact confidence values for each class. In contrast, Google's FER categorizes its output using five distinct likelihood descriptors, (\mie, ``very\_unlikely'', ``unlikely'', ``possible'', ``likely'' and ``very\_likely'')  offering a different approach to presenting its analysis. This approach seems to be an attempt to obscure the confidence levels of the model's output. The central concept behind such quantization is to reduce the granularity of the confidence values returned by the API, thereby making it more challenging for potential attackers to discern intricate details about the model's internal processes. We mimicked Google's approach and assess the effectiveness of quantization by applying it to other APIs. Specifically, We achieve this by discretizing the confidence values into intervals and substituting the confidence value with the midpoint of the corresponding interval (\meg, $[0.5, 0.7)$ is represented as $0.6$). We manually reduce the granularity of the returned confidence information to examine its impact on the performance of the Model Extraction (ME) attack.

\begin{table}[!ht]\small
\setlength\tabcolsep{5pt}
\renewcommand{\arraystretch}{1.2}
    \centering
   
    \begin{tabular}{c|cccc}
    
     \multirow{2}{*}{API}   & \multicolumn{4}{c}{Dataset}\\
    \cline{2-5}
      &KDEF&RAFDB&EXPW&FER+\\
        \hline
\hline
Amazon&-1.54$\pm$0.1	&-2.76$\pm$0.11&	-1.98$\pm$0.14&	-0.45$\pm$0.03\\
\facepp&+0.43$\pm$0.08	&-0.76$\pm$0.18&	+0.31$\pm$0.12&	-0.98$\pm$0.08\\
    \end{tabular}
     \caption{Difference of attack fidelity w/ and w/o quantization augmentation.}\label{tbl:quanti}
\vspace{-6pt}
\end{table}

Table\mref{tbl:quanti} presents the analysis results. When applied to the Amazon API, quantization resulted in an average decrease of 1.68\% in the attack fidelity. In contrast, the \facepp API exhibits more mixed results in mitigating ME attacks.

Overall, the manual reduction of granularity in the returned confidence information has a somewhat detrimental effect on the results of the ME attack. However, it is not a sufficiently effective defense strategy against model extraction.

% Quantization is a technique that aims to blur the information provided by a model's output, rather than reducing the precision of the outputs themselves. The main idea is to limit the granularity of the confidence values returned through the API, making it more difficult for attackers to gain detailed information about the model's inner workings. This, in turn, can help prevent overfitting of the extracted model.

% Two approaches to quantization were considered:

% \begin{itemize}[noitemsep]
% \item Interval-based quantization, where a single point is chosen to represent all the data within an interval. This can be achieved by binning the confidence values into discrete intervals and then returning the midpoint of the corresponding interval.
% \item Label-based quantization, where only the predicted class label is returned, discarding the confidence value altogether.
% \end{itemize}

% To evaluate the effectiveness of these quantization methods, we will conduct experiments using various quantization levels and compare the extracted model's performance to the original model. 

\begin{mtbox}{\small Observation\,10}{\small
Quantization of query responses mitigates ME attacks to a limited extent but is insufficient.}
\end{mtbox}

% \section{ Potential defenses }
% (e.g., quantization)
% THIEVES ON SESAME STREET! MODEL EXTRACTION OF BERT-BASED APIS

\vspace{-3pt}
\section{A retrospective study}

To comprehend the evolution of ME vulnerability over the years, we integrate a longitudinal dataset into \system and conduct a retrospective study on the ME vulnerability of leading MLaaS APIs, spanning the period from 2019 until now. The purpose of this study is to understand the evolving security risks associated with ME attacks. 
%offering stakeholders a deeper insight into the threat landscape surrounding MLaaS APIs. 
% This knowledge can, in turn, aid in the development of more effective strategies to counter ME attacks.

% In an effort to comprehend the progression of ME vulnerability, we undertake a comprehensive analysis of mainstream MLaaS APIs, covering the period from 2019 until now. This study is designed to illustrate the evolving security risks associated with ME attacks and thus provide stakeholders with an understanding of the threat landscape for MLaaS APIs. This knowledge can facilitate the design of more effective strategies to counter ME attacks.

% \begin{table}[!ht]
% \setlength\tabcolsep{4pt}
% \renewcommand{\arraystretch}{1.2}
%     \centering
%     \caption{The generalizability of different datasets and APIs of FER}
%     \label{tbl:trans_fer}
%     \begin{tabular}{c|c|cccc}

%           \multirow{2}{*}{Valid Dataset} &  \multirow{2}{*}{API} & 
%     \multicolumn{4}{c}{Origin Dataset} \\
%    \cline{3-6}
% & &   KDEF&   RAFDB & EXPW & FER+       \\

\subsection{Study setting}
\label{sec:data_pre}

In this retrospective study, we employ HAPI\mcite{chen2022hapi}, a longitudinal compilation containing 1,761,417 queries submitted to commercial MLaaS APIs including Amazon, Google, IBM and Microsoft. The dataset spans from 2020 to 2022 and covers a range of tasks, such as image tagging, speech recognition, and text mining. Each data point comprises a query input along with the MLaaS's response (\mie, prediction, annotation, and corresponding confidence scores).

Among the range of tasks covered by the HAPI dataset, we focus on two representative tasks due to the inherent constraints of ME: FER (representing computer vision tasks) and NLU (representing natural language processing tasks). Note that the responses of all the APIs include confidence information for all the classes, with the notable exception of Google's API (details in Table\mref{tbl:api}).

\begin{table}[htbp]\small
\setlength\tabcolsep{10pt}
\renewcommand{\arraystretch}{1.15}
    \centering
\begin{threeparttable} 
    \begin{tabular}{c|c|ccc}
     
        Task&API & HAPI\tnote{1}&\makecell{Current\tnote{2}}&\makecell{\#\,Classes}  \\ % & \makecell{Full Confidence\\ Information\tnote{3}}
        \hline
                \hline
       \multirow{4}{*}{FER}& Amazon& \ding{55} &\ding{51}& 8 \\
        &Microsoft\tnote{3}& \ding{51} &\ding{55}\tnote{3} & 7 \\
        &Face++ &\ding{51} &\ding{51}& 7 \\
        &Google  & \ding{51} &\ding{51} & 7   \\
        \hline
         \multirow{2}{*}{NLU} &Amazon & \ding{51} & \ding{51} & 4 \\
        &Microsoft& \ding{55} &\ding{51} & 3\\
    \end{tabular}
         \caption{Details of MLaaS APIs.\label{tbl:api}}
\vspace{-6pt}
            \begin{tablenotes}
                \footnotesize
                \item[1] ``HAPI'' represents the data collected by HAPI during 2020-2022 with only the highest confidence scores.
                \item[2] ``Current'' represents the data collected by us in 2023 with full confidence scores across all classes, except for Google API which only returns the highest confidence scores.
                \item[3] Microsoft has sunset its FER API for emotion prediction to mitigate potential misuse that subject people to stereotyping, discrimination, or unfair denial of services.
                % ... 其他注释 ...
            \end{tablenotes}

    \end{threeparttable}
\end{table}

Due to the unique structures of the HAPI dataset, we face a particular challenge: it provides only the highest confidence score among all the classes, without any information about the remaining classes. To address this challenge, we impute the confidence scores of the remaining classes. Specifically, let the prediction about an input $x$ be a probability simplex $[f_v^0(x), \ldots, f_v^{m-1}(x)]$ over $m$ classes, with each element corresponding to one distinct class. Recall that the API only outputs the highest confidence score, $f^i_v(x)$, where $i = \arg\max_j f_v^j(x)$.  Following the maximum-entropy principle\mcite{jaynes1957information}, which suggests choosing a distribution that retains the most significant degree of uncertainty (or entropy) while adhering to known constraints, we thus assign the confidence scores as:
% take into account the output structures of the API, denoted as $f_v(x)$. This output encompasses $m$ confidence scores, with each corresponding to a distinct class: $[f_v^0(x), f_v^1(x), \ldots, f_v^{m-1}(x)]$. However, as mentioned earlier, HAPI exclusively outputs the highest confidence score, represented as $f_v^\mathrm{hapi}(x)$, which pertains to class $i$.
% To interpolate the missing components  $f_v^j(x)$, we draw inspiration from the Maximum Entropy Principle \cite{jaynes1957information}. This principle suggests choosing a distribution that retains the most significant degree of uncertainty (or entropy) while adhering to known constraints. Specifically, we assign the same value to the missing components:
\begin{equation}
\label{eq:datapre}
f_v^j(x) = 
\begin{cases} 
f_v^\mathrm{hapi}(x) & \text{if } j=i \\
\frac{1 - f_v^\mathrm{hapi}(x)}{m - 1} & \text{if } j \neq i
\end{cases}
\end{equation}
That is, the class with the highest confidence score is assigned the score given by HAPI, while the remaining classes receive an equal assignment \((1 - f_v^\mathrm{hapi}(x))/(m - 1)\). Note that the sum of confidence scores across all classes equals to 1, ensuring a valid probability distribution.
% \begin{equation}
% \label{eq:datapre}
% f^lb_v(x) = f_v^{hapi}(x)
% \\
%  f^m_v(x)  = \mathds{1 - f_v^{hapi}(x)}{number\; of\;  classes - 1}
% \end{equation}

% \begin{equation}
% \frac{\sum_{(x, y) \in \gD}\mathds{1}_{f_t(x) = f_p(x)}}{|\gD_\mathrm{evl}|}
% \end{equation}

Also note that in the NLU task, as HAPI only records the highest confidence score between ``positive'' and ``negative'', we expand the outputs to three classes: ``positive'', ``negative'', and ``mixed'' in the following study.

\begin{figure}[!t]
    \centering
    \includegraphics[width=\linewidth]{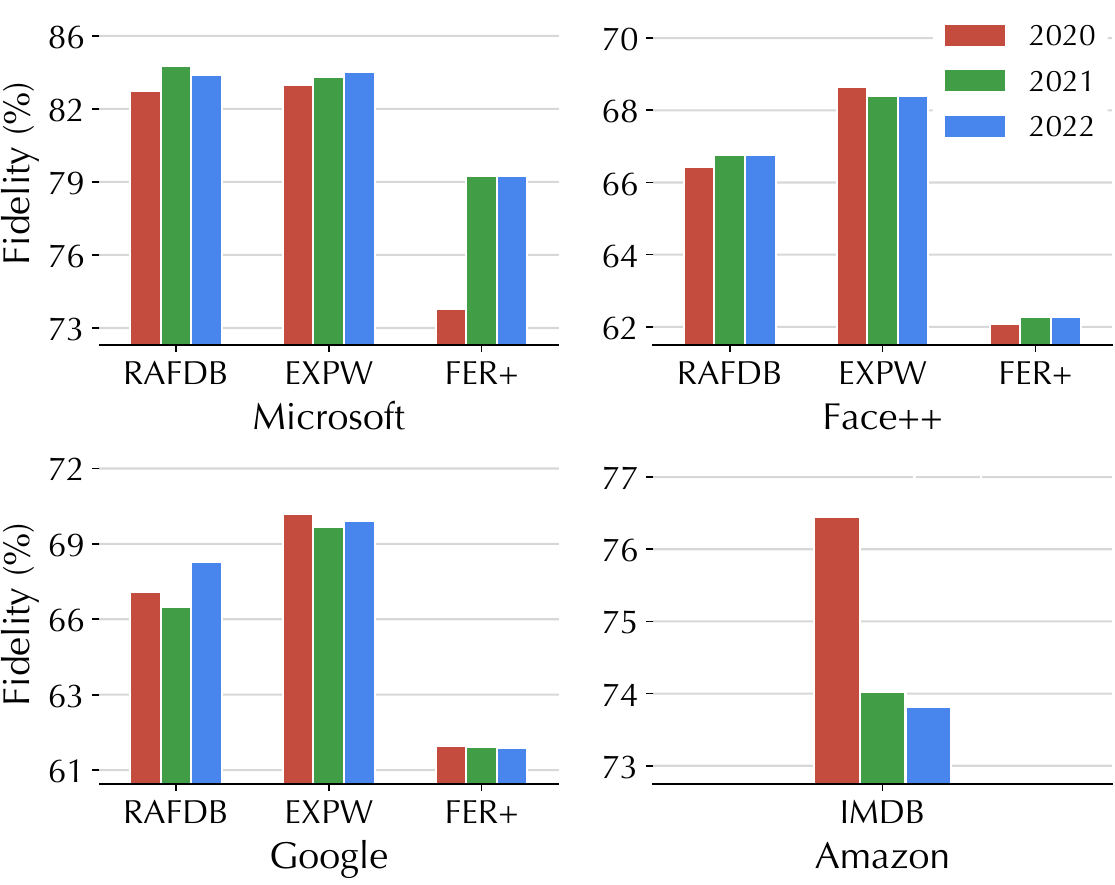}
    \caption{A retrospective study of ME attack vulnerability across different datasets and APIs.}
    \label{fig:retro}
\end{figure}

\subsection{Results}

Our study uncovers interesting trends in the evolving vulnerability of MLaaS APIs to ME attacks. Next, we present our findings in both FER and NLU tasks.

% Drawing insights from Figure 1, we observe the following noteworthy findings:

\subsubsection{FER}

Figure\mref{fig:retro} measures the attack fidelity across various APIs and datasets from the year 2020 to 2022. Compared with other platforms, there is a general upward trend for the attack fidelity against the Microsoft API across all three datasets, with a significant shift from 2020 to 2021 on the FER+ dataset. To understand the reason behind this shift, we further conduct a detailed analysis, summarizing the variations in the outcomes of ME attacks and the output data of original APIs in Table\mref{tbl:retro_fer}.

% \jc{update this section} Analyzing Figure \mref{fig:retro}, it is evident that the Microsoft API demonstrates a consistent upward trend across all three datasets from 2020 to 2021, with a particularly notable surge in FER+. We tried to find the reason for this shift in the API's returned data. The variations in model extraction over this two-year span, along with data from the original APIs, are detailed in Table \ref{tbl:retro_fer}.

\begin{table}[!ht]\small
\setlength\tabcolsep{2pt}
\renewcommand{\arraystretch}{1.2}
    \centering
    \begin{tabular}{c|c|c|c|c}
     \multirow{2}{*}{Dataset} & ME Attack &\multicolumn{3}{c}{API Ouput} \\
     \cline{2-5}
&Fidelity&Accuracy&Predicted Class&Avg. Confidence\\
     
    \hline
    \hline
FER+&\cellcolor{Red}+5.93\%&\cellcolor{Red}+3.03\%&\cellcolor{Red}96.07\%&\cellcolor{Red}0.0169\\
      RAFDB&+1.10\%& -0.01\%&99.76\%&0.0022\\
      EXPW&+0.38\%& +0.03\%&99.44\%&0.0059\\
      % FER+&\cellcolor{Red}+5.93\%&\cellcolor{Red}81.39\% +3.03\%&\cellcolor{Red}96.07\%&\cellcolor{Red}0.0169\\
      % RAFDB&+1.10\%& 71.68\% -0.01\%&99.76\%&0.0022\\
      % EXPW&+0.38\%&72.79\% +0.03\%&99.44\%&0.0059\\
    \end{tabular}
    \caption{Change of the Microsoft API from 2020 to 2021 in (i) the fidelity of the ME attack, (ii) the overall accuracy, (iii) predicted classes, and (iv) average confidence scores of the API.}
    \label{tbl:retro_fer}
     \vspace{-8pt}
    
\end{table}

\vspace{-3pt}
When examining the API outputs with respect to the HAPI dataset, we find that the predictions in 2020 and 2021 overlap more than 99\% for RAFDB and EXPW, while this overlap is only around 96\% for FER+. It is important to note that these overlaps only match predicted labels and do not account for variations in confidence scores. When considering the average change in confidence scores, FER+ also exhibits much higher variance than other datasets. Finally, the overall accuracy of the Microsoft API in classifying FER+ also increases by 3.03\% from 2020 to 2021, while the accuracy for the other two datasets remains little changed.

% On assessing the Origin API's (HAPI data) outcomes, the \textbf{prediction overlaps} for RAFDB and EXPW were both above 99\%. For FER+, this overlap was 99.6\%. It is important to clarify that these overlap percentages pertain solely to the matching predicted labels and don't account for potential variations in confidence levels. 

% In fact, when considering shifts in \textbf{confidence levels}, both EXPW and FER+ exhibit significant discrepancies, with overlaps of just around 67\%. Though both RAFDB and FER+ underwent alterations in return confidence metrics, their ME attack outcomes diverge. FER+ saw a marked rise, EXPW had a modest boost, but RAFDB declined. 

The substantial changes in the Microsoft API's accuracy and confidence scores with respect to the FER+ and EXPW datasets suggest that the backend model may have undergone significant changes from 2020 to 2021. These alterations likely contribute to the varying fidelity of ME attacks. Specifically, the attack attains 5.93\% fidelity increase on FER+, with only minor increments of 1.10\% and 0.38\% on RAFDB and EXPW, respectively.

% Given the notable changes in platform accuracy and confidence observed in both the FER+ and EXPW datasets, it is evident that Microsoft's model was modified between 2020 and 2021. This likely contributed to the varying results in the ME attacks. 

% Specifically, the FER+ dataset observed a significant growth of 5.93\%, RAFDB and expw have with a minor increment of 1.10\% and 0.38\%, respectively.

While ME attacks appear to become both easier and simpler across different datasets, we believe such changes are mainly due to adjustments in the backend model of the Microsoft API (\meg incorporating new training data to fine-tune the model), rather than any alterations to its defenses against ME attacks. What is particularly interesting is that the significant improvements in model accuracy coincide with the increased attack fidelity. We speculate that after the update, the backend model becomes more aligned with the distribution of the FER+ dataset. As the ME attack specifically targets this dataset, it is easier for the attack to adapt to the backend model. In comparison, Google+EXPW shows an overlap of 90.83\% between 2020 and 2021 in HAPI, with a slight fidelity change of -0.47\%. All the other dataset-API combinations experience minimal changes, with overlaps consistently exceeding 99\%.

% ME attack seems to get easier and simpler at the same time between different datasets. However, we consider that the recent changes in Microsoft ME attack outcomes can be mainly due to the adjustments in the Microsoft model(\meg incorporate new training data to finetune the model), rather than any alterations to the API's defenses against ME attacks. What is more interesting is that dramatic improvements in model accuracy coincided with increased fidelity in model extraction. We consider that after the model update, it might align more closely with the distribution of FER+ dataset. Given that our ME attack specifically targets this dataset, it becomes easier for us to adapt to the updated model.

% For other services, Google+EXPW showed an overlap of 90.83\% between 2020-2021 in HAPI, with a fidelity change of -0.47\% for ME attacks. All other datasets and APIs combinations change very small, overlap is above 99\%.

In summary, it is evident that in the FER task, the ME vulnerability of various MLaaS APIs has evolved in the past few years. However, we do not observe clear patterns in such evolution and are therefore unable to conclude whether these APIs have strengthened defenses against ME attacks. We tend to believe that most MLaaS platforms have not implemented specific defenses, especially in the case of the Microsoft API, where it exhibits high vulnerability to ME attacks.

% In summary, it is clear that the vulnerability of different MLaaS APIs in FER has changed, but this change presents inconsistency between the datasets, and we are unable to conclude that they have strengthened their defenses against ME attacks during these three years. We tend to consider that they do not make special defenses against ME attacks, especially Microsoft APIs, their models show high fidelity under ME attacks. it is easy to be stolen by attackers.

\subsubsection{NLU} Due to the limitation of NLU datasets in the HAPI dataset, only the combination of Amazon and IMDB is available for analyzing the evolution of ME vulnerability. 
     % \vspace{-6pt}

\begin{table}[!ht]\small
\setlength\tabcolsep{3pt}
\renewcommand{\arraystretch}{1.25}
    \centering
   
       \begin{tabular}{c|c|c|c|c}
     \multirow{2}{*}{Year} & ME Attack &\multicolumn{3}{c}{API Ouput} \\
     \cline{2-5}
&Fidelity&Accuracy&Predicted Class&Avg. Confidence\\
    \hline
2020-2021&\cellcolor{Red}-2.43\%&\cellcolor{Red}-1.08\%&\cellcolor{Red}83.20\% &\cellcolor{Red}0.1682\\

2021-2022&-0.20\%&+0.12\%&99.41\%&0.0070\\

    \end{tabular}
     \caption{Change of the Amazon API from 2020 to 2022 in (i) the fidelity of the ME attack, (ii) the overall accuracy, (iii) predicted classes, and (iv) average confidence scores of the API.}
     \label{tbl:hapi-nlp}
     \vspace{-6pt}
\end{table}
     % \vspace{-6pt}

% record: 20799 total: 25000 percent 0.83196
% recordcon: 0 total: 25000 percent 0.0
% sum tensor(-2420.8447) avg:  tensor(-0.0968)
% record: 24853 total: 25000 percent 0.99412
% recordcon: 23867 total: 25000 percent 0.95468
% sum tensor(-18.2823) avg:  tensor(-0.0007)

As shown in Table\mref{tbl:hapi-nlp}, the attack fidelity against the Amazon API shows a declining trend over the three years, with a decrease of 2.43\% from 2020 to 2021. However, since this change is not statistically significant, it does not appear that the API has fortified its defenses against ME attacks in the intervening years.

Despite the growing popularity of MLaaS APIs, our study indicates a potential lack of investment in safeguarding backend models. This poses questions about the commitment of commercial MLaaS providers to addressing ME threats and implementing protective measures. Considering the ongoing expansion and increasing reliance on MLaaS, it becomes imperative for providers to prioritize security. We advocate for a proactive stance by MLaaS providers to bolster their APIs and fortify the protection of their backend models.

%% file: literature.tex
\section{Additional related work}
Since the concept of ME attacks against MLaaS APIs was first introduced in\mcite{tramer2016stealing}, which applies path-finding attacks to extract classification (\meg, decision trees) and regression (\meg, regression trees) models, over the years, a plethora of ME attacks have been proposed, training the piracy model to mimic the behavior of the victim model by leveraging data labeled by the victim model\mcite{tramer2016stealing, orekondy2019knockoff, krishna2019thieves, chandrasekaran2020exploring, pal2019framework, pal2020activethief, pengcheng2018query, shi2018active, juuti2019prada, jagielski2020high, yu2020cloudleak, Truong_2021_CVPR, carlini2021extracting, papernot2016transferability, papernot2016distillation, papernot2018sok, papernot2017practical}. % Our research goal is consistent with them.

Two key metrics, accuracy and fidelity, have been proposed in\mcite{jagielski2020high} to measure the performance of ME attacks. Fidelity is generally acknowledged as a more critical metric than accuracy for assessing ME attacks, which is also the main metric used in this study. In addition, we also introduce adversarial fidelity, another important metric to assess the agreement between the victim and piracy models with respect to adversarial examples, complementing fidelity and accuracy.

The number of queries is another important metric for ME attacks. Semi-supervised learning has been explored in\mcite{jagielski2020high} to reduce the number of queries. Active learning strategies have also been explored in\mcite{tramer2016stealing, chandrasekaran2020exploring, pal2019framework, pal2020activethief, pengcheng2018query, shi2018active}. For instance, in\mcite{pal2020activethief},  strategies such as uncertainty, $K$-center, and DeepFool-based active learning are employed to identify the most informative samples. While the experiments span both image and language-based tasks, they are conducted in controlled laboratory environments and not validated on MLaaS platforms. Other work\mcite{papernot2017practical, juuti2019prada, pengcheng2018query, yu2020cloudleak} leverages adversarial examples to optimize the number of queries or improve attack performance. In addition to these studies, which have focused primarily on classification tasks, \mcite{hu2021stealing} explores model extraction attacks on GANs.
This work examines whether the conclusions in controlled local environments hold for ME attacks against real-world MLaaS APIs, leading to a number of interesting findings that complement existing studies and provide new insights about the vulnerability of real-world MLaaS APIs.

% We seek to refine the experimental setups and systematically compare their effects across the same platform to derive deeper insights.

% ME attack has garnered significant attention in recent years, leading to the development of various techniques and countermeasures to protect machine learning models. In addition to the previously mentioned works, several other studies have contributed to the understanding of ME attacks and their implications. \cite{jagielski2020high} proposed a framework for quantifying the vulnerability of models to extraction attacks, demonstrating that certain model architectures are more susceptible to being extracted than others. They also highlighted the trade-offs between model vulnerability and performance.

% PRADA\cite{juuti2019prada} proposed a detection mechanism for Deep Neural Network (DNN) ME attacks. PRADA analyzes the distribution of consecutive API queries and raises an alarm when the distribution deviates from benign behavior. Another line of research has focused on watermarking techniques for neural networks as a means to detect unauthorized ME. 

% \cite{zhang2020pushing,hitaj2018have}. These techniques embed unique signatures into the target model's parameters, which can then be used to prove ownership in the event of a model theft. In conclusion, the growing body of research on ME attacks highlights the importance of understanding the vulnerabilities of machine learning models and developing effective countermeasures to protect intellectual property and user privacy.

%% file: Conclusion.tex
\section{Conclusion}
In this paper, we report a systematic study on the vulnerability of real-world machine-learning-as-a-service (MLaaS) APIs to model extraction (ME) attacks. The evaluation leads to a number of interesting findings that complement the existing studies conducted in controlled laboratory environments: \mct{i} Despite their striding progression over the years, leading MLaaS platforms continue to be highly susceptible to ME attacks. \mct{ii} The advances in ML techniques (\meg, optimizers) significantly enhance the adversary's capabilities. \mct{iii} Attack techniques (\meg, adversarial learning), proven to be effective in controlled environments, may not necessarily exhibit the same level of effectiveness against real-world MLaaS APIs. \mct{iv} The existing mitigation (\meg, output quantization) may weaken ME attacks but remain inadequately effective. These findings shed light on developing MLaaS platforms in a more secure manner.

generally improves fidelity but has a negative impact on adversarial fidelity, active learning provides a marginal enhancement, and surprisingly, adversarial learning may even negatively impact ME attacks. \mct{v} The transferability of extracted piracy models depends more on the datasets, rather than other factors such as APIs, models, or ME attacks. \mct{vi} API output quantization may weaken the effect of ME attacks, yet it remains inadequately effective.

%% file: appendix.tex
\appendix

\section{More Details}

\subsection{Notations}

\begin{table}[!ht]\small
\renewcommand{\arraystretch}{1.25}
    \centering
    \label{tbl:opti}
    \begin{tabular}{c|c}
    Notation & Definition\\
    \hline
    \hline
$f_v$, $f_v^\mathrm{real}$, $f_v^\mathrm{hapi}$ & generic, online, HAPI victim model\\
$f_p$& piracy model\\
$\gX$& clean labeled data\\
$\gU$& clean unlabeled data\\
$\gA$& adversarial data\\
    \end{tabular}
     % \begin{tablenotes}
     %        \small
     %        \item[1] ``HAPI'' only outputs the highest confidence score $f_v^\mathrm{hapi} \leftarrow \max (f_v^\mathrm{real})$
     %      \end{tablenotes}
\caption{Notations and symbols.}
\end{table}

\subsection{Datasets}

\begin{table}[!ht]\small
\renewcommand{\arraystretch}{1.2}
    \centering
    \begin{tabular}{c|m{1cm}<{\centering}|m{1.1cm}<{\centering}|m{4cm}}
Task&Dataset &Size& Description\\
    \hline
    \hline
    \multirow{6}{*}{\makecell{F\\E\\R}} & KDEF& 562*762&The exaggerated facial expressions made by subjects\\
    &RAFDB&100*100&Great-diverse facial images downloaded from the Internet\\
    &EXPW&224*224&Wild human facial expressions captured at social events\\
    &FER+&48*48 &Black and white human face expressions\\
    \hline
    
    \multirow{4}{*}{\makecell{N\\L\\U}}&IMDB&234 words& Binary sentiment classification dataset, highly polar movie reviews collected from the internet movie database (IMDB) \\
    &Yelp&140 words& Information from user reviews of different restaurants on Yelp. Ratings can be from 1-5, not a polarized dataset\\

    \end{tabular}
        \caption{Details of datasets.   \label{tbl:dataset}}
\end{table}

\subsection{Local experiments}

To validate the correctness of our experimental setting, we perform local experiments using the CIFAR10 dataset. We train a victim model using the ground-truth labels and use the same strategy in\msec{sec:advanceattack} to perform ME attack against it. The results are summarized in Table\mref{tbl:cfmixmatch}, \mref{tbl:cfactive}, and \mref{tbl:cfadv}.

\begin{table}[!ht]\small
\setlength\tabcolsep{3pt}
\renewcommand{\arraystretch}{1.2}
    \centering
     \begin{tabular}{c|ccccccc}
     \multirow{2}{*}{Attack}&  \multicolumn{6}{c}{Query Budget (\#\,Batches)} \\
   \cline{2-7}
&    4 & 8 & 16 & 32 & 64 & Full   \\
\hline
\hline

  Basic & 35.05	&50.41	&49.46	&71.98	&73.92 & \multirow{2}{*}{92.88}  \\
 MixMatch &49.35	&58.68	&79.21&	79.22&	83.23& \\

    \end{tabular}
     \caption{Fidelity of basic and MixMatch-based ME attacks (with batch size fixed as 64). In the case of MixMatch, each labeled batch is paired with an equal-sized unlabeled batch.}
  
\label{tbl:cfmixmatch}
\end{table}

\begin{table}[!ht]\small
\setlength\tabcolsep{4pt}
\renewcommand{\arraystretch}{1.25}
    \centering
    \begin{tabular}{c|cccccc}

  \multirow{2}{*}{Attack} &   \multicolumn{6}{c}{Query Budget (\#\,Batches)} \\
   \cline{2-7}
  &8	&16	&32	&64	&128 & Full   \\
\hline
\hline

 Basic& 47.85	&48.12&	68.84	&76.63&	83.12 & \multirow{2}{*}{\makecell{92.88}} \\

ActiveThief  &52.01	&58.57	&74.22&	80.53&	86.06		&  \\

    \end{tabular}
    \caption{Fidelity of basic and active learning-based ME attacks (with batch size fixed as 64).}
    \label{tbl:cfactive}
\end{table}

\begin{table}[!ht]\small
\setlength\tabcolsep{3pt}
\renewcommand{\arraystretch}{1.25}

    \centering
 
    \begin{tabular}{c|ccc}

    \multirow{2}{*}{\makecell{Query\\Batches}} & 
    \multicolumn{3}{c}{Attack} \\
       \cline{2-4}
  &   Basic & PGD & CW      \\
\hline
\hline
8&33.94&	36.87&	35.67\\

16&41.5&45.16&46.76\\

32&66.78&69.81&71.2\\

%        \multirow{3}{*}{Amazon} & \multirow{3}{*}{KDEF} & 8 & \cellcolor{Red}62.33(44.44) &53.99(37.85) & 55.56(44.79)  \\
% && 16 & \cellcolor{Red}64.06(45.66) &56.60(38.54) & 55.73(39.58) \\
% && 32 & \cellcolor{Red}67.71(48.61) &62.15(41.15) & 64.06(44.97)  \\
% \hline

% \multirow{3}{*}{\facepp}& \multirow{3}{*}{KDEF} & 8 & 56.94(52.95) &52.43(46.01) & \cellcolor{Red}57.64(50.17)  \\
% && 16 & \cellcolor{Red}68.06(60.59) &60.07(54.17) & 62.67(57.12) \\
% && 32 & \cellcolor{Red}70.31(64.58) &64.58(59.03) & 66.15(61.98)  \\

% \hline

% \multirow{3}{*}{\facepp} &\multirow{3}{*}{RAFDB} & 8 & \cellcolor{Red}41.22(42.99) &33.84(39.43) & 36.27(39.96)  \\
% && 16 & \cellcolor{Red}47.41(45.31) &34.57(44.45) & 39.86(41.59) \\
% && 32 & \cellcolor{Red}46.51(48.67) &44.25(47.11) & 44.35(47.34) \\
    \end{tabular}
       \caption{Fidelity of basic and adversarial learning-based ME attacks (with batch size fixed as 64).}
        \label{tbl:cfadv}
\end{table}

%% file: main.bbl

\begin{thebibliography}{85}

%%% the Web address.

\ifx \showCODEN    \undefined \def \showCODEN     #1{\unskip}     \fi
\ifx \showDOI      \undefined \def \showDOI       #1{#1}\fi
\ifx \showISBNx    \undefined \def \showISBNx     #1{\unskip}     \fi
\ifx \showISBNxiii \undefined \def \showISBNxiii  #1{\unskip}     \fi
\ifx \showISSN     \undefined \def \showISSN      #1{\unskip}     \fi
\ifx \showLCCN     \undefined \def \showLCCN      #1{\unskip}     \fi
\ifx \shownote     \undefined \def \shownote      #1{#1}          \fi
\ifx \showarticletitle \undefined \def \showarticletitle #1{#1}   \fi
\ifx \showURL      \undefined \def \showURL       {\relax}        \fi
% invisible to TeX
\providecommand\bibfield[2]{#2}
\providecommand\bibinfo[2]{#2}
\providecommand\natexlab[1]{#1}
\providecommand\showeprint[2][]{arXiv:#2}

\bibitem[imd({[n.\,d.]})]%
        {imdb}
 \bibinfo{year}{[n.\,d.]}\natexlab{}.
\newblock \bibinfo{title}{IMDb Datasets}.
\newblock \bibinfo{howpublished}{\url{https://www.imdb.com/interfaces/}}.
\newblock
\newblock
\shownote{Accessed: 2023-03-16}.


\bibitem[yel({[n.\,d.]})]%
        {yelp}
 \bibinfo{year}{[n.\,d.]}\natexlab{}.
\newblock \bibinfo{title}{Yelp Datasets}.
\newblock \bibinfo{howpublished}{\url{https://www.yelp.com/dataset}}.
\newblock
\newblock
\shownote{Accessed: 2023-07-16}.


\bibitem[A{\"\i}vodji et~al\mbox{.}(2020)]%
        {aivodji2020model}
\bibfield{author}{\bibinfo{person}{Ulrich A{\"\i}vodji}, \bibinfo{person}{Alexandre Bolot}, {and} \bibinfo{person}{S{\'e}bastien Gambs}.} \bibinfo{year}{2020}\natexlab{}.
\newblock \showarticletitle{Model extraction from counterfactual explanations}.
\newblock \bibinfo{journal}{\emph{ArXiv e-prints}} (\bibinfo{year}{2020}).
\newblock


\bibitem[Amazon({[n.\,d.]})]%
        {amazon-aws}
\bibfield{author}{\bibinfo{person}{Amazon}.} \bibinfo{year}{[n.\,d.]}\natexlab{}.
\newblock \bibinfo{title}{AWS Rekognition documentation}.
\newblock \bibinfo{howpublished}{\url{https://docs.aws.amazon.com/ rekognition/latest/dg/what-is.html}}.
\newblock


\bibitem[Barsoum et~al\mbox{.}(2016)]%
        {BarsoumICMI2016}
\bibfield{author}{\bibinfo{person}{Emad Barsoum}, \bibinfo{person}{Cha Zhang}, \bibinfo{person}{Cristian Canton~Ferrer}, {and} \bibinfo{person}{Zhengyou Zhang}.} \bibinfo{year}{2016}\natexlab{}.
\newblock \showarticletitle{Training Deep Networks for Facial Expression Recognition with Crowd-Sourced Label Distribution}. In \bibinfo{booktitle}{\emph{Proceedings of ACM International Conference on Multimodal Interaction (ICMI)}}.
\newblock


\bibitem[Berthelot et~al\mbox{.}(2019)]%
        {mixmatch}
\bibfield{author}{\bibinfo{person}{David Berthelot}, \bibinfo{person}{Nicholas Carlini}, \bibinfo{person}{Ian~J. Goodfellow}, \bibinfo{person}{Nicolas Papernot}, \bibinfo{person}{Avital Oliver}, {and} \bibinfo{person}{Colin Raffel}.} \bibinfo{year}{2019}\natexlab{}.
\newblock \showarticletitle{MixMatch: {A} Holistic Approach to Semi-Supervised Learning}.
\newblock \bibinfo{journal}{\emph{ArXiv e-prints}} (\bibinfo{year}{2019}).
\newblock


\bibitem[Cao et~al\mbox{.}(2018)]%
        {cao2018vggface2}
\bibfield{author}{\bibinfo{person}{Qiong Cao}, \bibinfo{person}{Li Shen}, \bibinfo{person}{Weidi Xie}, \bibinfo{person}{Omkar~M Parkhi}, {and} \bibinfo{person}{Andrew Zisserman}.} \bibinfo{year}{2018}\natexlab{}.
\newblock \showarticletitle{Vggface2: A dataset for recognizing faces across pose and age}. In \bibinfo{booktitle}{\emph{Proceedings of the IEEE International Conference on Automatic Face \& Gesture Recognition (FG)}}.
\newblock


\bibitem[Carlini et~al\mbox{.}(2021)]%
        {carlini2021extracting}
\bibfield{author}{\bibinfo{person}{Nicholas Carlini}, \bibinfo{person}{Florian Tramer}, \bibinfo{person}{Eric Wallace}, \bibinfo{person}{Matthew Jagielski}, \bibinfo{person}{Ariel Herbert-Voss}, \bibinfo{person}{Katherine Lee}, \bibinfo{person}{Adam Roberts}, \bibinfo{person}{Tom~B Brown}, \bibinfo{person}{Dawn Song}, \bibinfo{person}{Ulfar Erlingsson}, {et~al\mbox{.}}} \bibinfo{year}{2021}\natexlab{}.
\newblock \showarticletitle{Extracting Training Data from Large Language Models.}. In \bibinfo{booktitle}{\emph{Proceedings of USENIX Security Symposium (SEC)}}.
\newblock


\bibitem[Carlini and Wagner(2017)]%
        {carlini2017towards}
\bibfield{author}{\bibinfo{person}{Nicholas Carlini} {and} \bibinfo{person}{David Wagner}.} \bibinfo{year}{2017}\natexlab{}.
\newblock \showarticletitle{Towards evaluating the robustness of neural networks}. In \bibinfo{booktitle}{\emph{Proceedings of IEEE Symposium on Security and Privacy (S\&P)}}.
\newblock


\bibitem[Chandrasekaran et~al\mbox{.}(2020)]%
        {chandrasekaran2020exploring}
\bibfield{author}{\bibinfo{person}{Varun Chandrasekaran}, \bibinfo{person}{Kamalika Chaudhuri}, \bibinfo{person}{Irene Giacomelli}, \bibinfo{person}{Somesh Jha}, {and} \bibinfo{person}{Songbai Yan}.} \bibinfo{year}{2020}\natexlab{}.
\newblock \showarticletitle{Exploring connections between active learning and model extraction}. In \bibinfo{booktitle}{\emph{Proceedings of USENIX Security Symposium (SEC)}}.
\newblock


\bibitem[Chen et~al\mbox{.}(2022a)]%
        {chen2022hapi}
\bibfield{author}{\bibinfo{person}{Lingjiao Chen}, \bibinfo{person}{Zhihua Jin}, \bibinfo{person}{Sabri Eyuboglu}, \bibinfo{person}{Christopher R{\'e}}, \bibinfo{person}{Matei Zaharia}, {and} \bibinfo{person}{James Zou}.} \bibinfo{year}{2022}\natexlab{a}.
\newblock \showarticletitle{HAPI: A Large-scale Longitudinal Dataset of Commercial ML API Predictions}.
\newblock \bibinfo{journal}{\emph{ArXiv e-prints}} (\bibinfo{year}{2022}).
\newblock


\bibitem[Chen et~al\mbox{.}(2023a)]%
        {lion}
\bibfield{author}{\bibinfo{person}{Xiangning Chen}, \bibinfo{person}{Chen Liang}, \bibinfo{person}{Da Huang}, \bibinfo{person}{Esteban Real}, \bibinfo{person}{Kaiyuan Wang}, \bibinfo{person}{Yao Liu}, \bibinfo{person}{Hieu Pham}, \bibinfo{person}{Xuanyi Dong}, \bibinfo{person}{Thang Luong}, \bibinfo{person}{Cho-Jui Hsieh}, \bibinfo{person}{Yifeng Lu}, {and} \bibinfo{person}{Quoc~V. Le}.} \bibinfo{year}{2023}\natexlab{a}.
\newblock \showarticletitle{Symbolic Discovery of Optimization Algorithms}.
\newblock \bibinfo{journal}{\emph{ArXiv e-prints}} (\bibinfo{year}{2023}).
\newblock


\bibitem[Chen et~al\mbox{.}(2022b)]%
        {chen2022teacher}
\bibfield{author}{\bibinfo{person}{Yufei Chen}, \bibinfo{person}{Chao Shen}, \bibinfo{person}{Cong Wang}, {and} \bibinfo{person}{Yang Zhang}.} \bibinfo{year}{2022}\natexlab{b}.
\newblock \showarticletitle{Teacher model fingerprinting attacks against transfer learning}. In \bibinfo{booktitle}{\emph{Proceedings of USENIX Security Symposium (SEC)}}.
\newblock


\bibitem[Chen et~al\mbox{.}(2023b)]%
        {chen2023dark}
\bibfield{author}{\bibinfo{person}{Ziheng Chen}, \bibinfo{person}{Fabrizio Silvestri}, \bibinfo{person}{Jia Wang}, \bibinfo{person}{Yongfeng Zhang}, {and} \bibinfo{person}{Gabriele Tolomei}.} \bibinfo{year}{2023}\natexlab{b}.
\newblock \showarticletitle{The dark side of explanations: Poisoning recommender systems with counterfactual examples}. In \bibinfo{booktitle}{\emph{Proceedings of the 46th International ACM SIGIR conference on Research and Development in Information Retrieval}}. \bibinfo{pages}{2426--2430}.
\newblock


\bibitem[Correia-Silva et~al\mbox{.}(2018)]%
        {correia2018copycat}
\bibfield{author}{\bibinfo{person}{Jacson~Rodrigues Correia-Silva}, \bibinfo{person}{Rodrigo~F Berriel}, \bibinfo{person}{Claudine Badue}, \bibinfo{person}{Alberto~F de Souza}, {and} \bibinfo{person}{Thiago Oliveira-Santos}.} \bibinfo{year}{2018}\natexlab{}.
\newblock \showarticletitle{Copycat cnn: Stealing knowledge by persuading confession with random non-labeled data}. In \bibinfo{booktitle}{\emph{Proceedings of the International Joint Conference on Neural Networks (IJCNN)}}.
\newblock


\bibitem[Devlin et~al\mbox{.}(2018)]%
        {devlin2018bert}
\bibfield{author}{\bibinfo{person}{Jacob Devlin}, \bibinfo{person}{Ming-Wei Chang}, \bibinfo{person}{Kenton Lee}, {and} \bibinfo{person}{Kristina Toutanova}.} \bibinfo{year}{2018}\natexlab{}.
\newblock \showarticletitle{Bert: Pre-training of deep bidirectional transformers for language understanding}.
\newblock \bibinfo{journal}{\emph{ArXiv e-prints}} (\bibinfo{year}{2018}).
\newblock


\bibitem[{Dosovitskiy} et~al\mbox{.}(2020)]%
        {vit}
\bibfield{author}{\bibinfo{person}{Alexey {Dosovitskiy}}, \bibinfo{person}{Lucas {Beyer}}, \bibinfo{person}{Alexander {Kolesnikov}}, \bibinfo{person}{Dirk {Weissenborn}}, \bibinfo{person}{Xiaohua {Zhai}}, \bibinfo{person}{Thomas {Unterthiner}}, \bibinfo{person}{Mostafa {Dehghani}}, \bibinfo{person}{Matthias {Minderer}}, \bibinfo{person}{Georg {Heigold}}, \bibinfo{person}{Sylvain {Gelly}}, \bibinfo{person}{Jakob {Uszkoreit}}, {and} \bibinfo{person}{Neil {Houlsby}}.} \bibinfo{year}{2020}\natexlab{}.
\newblock \showarticletitle{An Image is Worth 16x16 Words: Transformers for Image Recognition at Scale}. In \bibinfo{booktitle}{\emph{Proceedings of International Conference on Learning Representations (ICLR)}}.
\newblock


\bibitem[Feng et~al\mbox{.}(2021)]%
        {feng2021live}
\bibfield{author}{\bibinfo{person}{Xianglong Feng}, \bibinfo{person}{Weitian Li}, {and} \bibinfo{person}{Sheng Wei}.} \bibinfo{year}{2021}\natexlab{}.
\newblock \showarticletitle{LiveROI: Region of Interest Analysis for Viewport Prediction in Live Mobile Virtual Reality Streaming}. In \bibinfo{booktitle}{\emph{Proceedings of the ACM Multimedia Systems Conference (MMSys)}}.
\newblock


\bibitem[Gong et~al\mbox{.}(2021)]%
        {gong2021inversenet}
\bibfield{author}{\bibinfo{person}{Xueluan Gong}, \bibinfo{person}{Yanjiao Chen}, \bibinfo{person}{Wenbin Yang}, \bibinfo{person}{Guanghao Mei}, {and} \bibinfo{person}{Qian Wang}.} \bibinfo{year}{2021}\natexlab{}.
\newblock \showarticletitle{InverseNet: Augmenting Model Extraction Attacks with Training Data Inversion.}. In \bibinfo{booktitle}{\emph{Proceedings of International Joint Conference on Artificial Intelligence (IJCAI)}}.
\newblock


\bibitem[Gou et~al\mbox{.}(2021)]%
        {gou2021knowledge}
\bibfield{author}{\bibinfo{person}{Jianping Gou}, \bibinfo{person}{Baosheng Yu}, \bibinfo{person}{Stephen~J Maybank}, {and} \bibinfo{person}{Dacheng Tao}.} \bibinfo{year}{2021}\natexlab{}.
\newblock \showarticletitle{Knowledge distillation: A survey}.
\newblock \bibinfo{journal}{\emph{International Journal of Computer Vision}}  \bibinfo{volume}{129} (\bibinfo{year}{2021}), \bibinfo{pages}{1789--1819}.
\newblock


\bibitem[{He} et~al\mbox{.}(2015)]%
        {he2015deep}
\bibfield{author}{\bibinfo{person}{Kaiming {He}}, \bibinfo{person}{Xiangyu {Zhang}}, \bibinfo{person}{Shaoqing {Ren}}, {and} \bibinfo{person}{Jian {Sun}}.} \bibinfo{year}{2015}\natexlab{}.
\newblock \showarticletitle{Deep Residual Learning for Image Recognition}. In \bibinfo{booktitle}{\emph{Proceedings of IEEE Conference on Computer Vision and Pattern Recognition (CVPR)}}.
\newblock


\bibitem[He et~al\mbox{.}(2016)]%
        {he2016deep}
\bibfield{author}{\bibinfo{person}{Kaiming He}, \bibinfo{person}{Xiangyu Zhang}, \bibinfo{person}{Shaoqing Ren}, {and} \bibinfo{person}{Jian Sun}.} \bibinfo{year}{2016}\natexlab{}.
\newblock \showarticletitle{Deep residual learning for image recognition}. In \bibinfo{booktitle}{\emph{Proceedings of IEEE Conference on Computer Vision and Pattern Recognition (CVPR)}}.
\newblock


\bibitem[Hinton et~al\mbox{.}(2015)]%
        {knowdiss}
\bibfield{author}{\bibinfo{person}{Geoffrey Hinton}, \bibinfo{person}{Oriol Vinyals}, {and} \bibinfo{person}{Jeff Dean}.} \bibinfo{year}{2015}\natexlab{}.
\newblock \showarticletitle{Distilling the Knowledge in a Neural Network}.
\newblock \bibinfo{journal}{\emph{ArXiv e-prints}} (\bibinfo{year}{2015}).
\newblock


\bibitem[Hu and Pang(2021)]%
        {hu2021stealing}
\bibfield{author}{\bibinfo{person}{Hailong Hu} {and} \bibinfo{person}{Jun Pang}.} \bibinfo{year}{2021}\natexlab{}.
\newblock \showarticletitle{Stealing machine learning models: Attacks and countermeasures for generative adversarial networks}. In \bibinfo{booktitle}{\emph{Annual Computer Security Applications Conference}}. \bibinfo{pages}{1--16}.
\newblock


\bibitem[Huang et~al\mbox{.}({[n.\,d.]})]%
        {huangadversarial}
\bibfield{author}{\bibinfo{person}{Dong Huang}, \bibinfo{person}{Qingwen Bu}, \bibinfo{person}{Yuhao Qing}, \bibinfo{person}{Yichao Fu}, {and} \bibinfo{person}{Heming Cui}.} \bibinfo{year}{[n.\,d.]}\natexlab{}.
\newblock \showarticletitle{ADVERSARIAL FEATURE MAP PRUNING FOR BACK}.
\newblock  (\bibinfo{year}{[n.\,d.]}).
\newblock


\bibitem[Huang et~al\mbox{.}(2017)]%
        {huang2017densely}
\bibfield{author}{\bibinfo{person}{Gao Huang}, \bibinfo{person}{Zhuang Liu}, \bibinfo{person}{Laurens Van Der~Maaten}, {and} \bibinfo{person}{Kilian~Q Weinberger}.} \bibinfo{year}{2017}\natexlab{}.
\newblock \showarticletitle{Densely connected convolutional networks}. In \bibinfo{booktitle}{\emph{Proceedings of IEEE Conference on Computer Vision and Pattern Recognition (CVPR)}}.
\newblock


\bibitem[Huang et~al\mbox{.}(2018)]%
        {huang2018densely}
\bibfield{author}{\bibinfo{person}{Gao Huang}, \bibinfo{person}{Zhuang Liu}, \bibinfo{person}{Laurens van~der Maaten}, {and} \bibinfo{person}{Kilian~Q. Weinberger}.} \bibinfo{year}{2018}\natexlab{}.
\newblock \showarticletitle{Densely Connected Convolutional Networks}.
\newblock \bibinfo{journal}{\emph{ArXiv e-prints}} (\bibinfo{year}{2018}).
\newblock


\bibitem[Huang and Wang(2017)]%
        {huang2017like}
\bibfield{author}{\bibinfo{person}{Zehao Huang} {and} \bibinfo{person}{Naiyan Wang}.} \bibinfo{year}{2017}\natexlab{}.
\newblock \showarticletitle{Like what you like: Knowledge distill via neuron selectivity transfer}.
\newblock \bibinfo{journal}{\emph{ArXiv e-prints}} (\bibinfo{year}{2017}).
\newblock


\bibitem[Jagielski et~al\mbox{.}(2020)]%
        {jagielski2020high}
\bibfield{author}{\bibinfo{person}{Matthew Jagielski}, \bibinfo{person}{Nicholas Carlini}, \bibinfo{person}{David Berthelot}, \bibinfo{person}{Alex Kurakin}, {and} \bibinfo{person}{Nicolas Papernot}.} \bibinfo{year}{2020}\natexlab{}.
\newblock \showarticletitle{High accuracy and high fidelity extraction of neural networks}. In \bibinfo{booktitle}{\emph{Proceedings of USENIX Security Symposium (SEC)}}.
\newblock


\bibitem[Jaynes(1957)]%
        {jaynes1957information}
\bibfield{author}{\bibinfo{person}{Edwin~T Jaynes}.} \bibinfo{year}{1957}\natexlab{}.
\newblock \showarticletitle{Information theory and statistical mechanics}.
\newblock \bibinfo{journal}{\emph{Physical review}} \bibinfo{volume}{106}, \bibinfo{number}{4} (\bibinfo{year}{1957}), \bibinfo{pages}{620}.
\newblock


\bibitem[Juuti et~al\mbox{.}(2019)]%
        {juuti2019prada}
\bibfield{author}{\bibinfo{person}{Mika Juuti}, \bibinfo{person}{Sebastian Szyller}, \bibinfo{person}{Samuel Marchal}, {and} \bibinfo{person}{N Asokan}.} \bibinfo{year}{2019}\natexlab{}.
\newblock \showarticletitle{PRADA: protecting against DNN model stealing attacks}. In \bibinfo{booktitle}{\emph{Proceedings of IEEE European Symposium on Security and Privacy (Euro S\&P)}}.
\newblock


\bibitem[Kingma and Ba(2017)]%
        {adam}
\bibfield{author}{\bibinfo{person}{Diederik~P. Kingma} {and} \bibinfo{person}{Jimmy Ba}.} \bibinfo{year}{2017}\natexlab{}.
\newblock \showarticletitle{Adam: A Method for Stochastic Optimization}.
\newblock \bibinfo{journal}{\emph{ArXiv e-prints}} (\bibinfo{year}{2017}).
\newblock


\bibitem[Krishna et~al\mbox{.}(2019)]%
        {krishna2019thieves}
\bibfield{author}{\bibinfo{person}{Kalpesh Krishna}, \bibinfo{person}{Gaurav~Singh Tomar}, \bibinfo{person}{Ankur~P Parikh}, \bibinfo{person}{Nicolas Papernot}, {and} \bibinfo{person}{Mohit Iyyer}.} \bibinfo{year}{2019}\natexlab{}.
\newblock \showarticletitle{Thieves on sesame street! model extraction of bert-based apis}.
\newblock \bibinfo{journal}{\emph{ArXiv e-prints}} (\bibinfo{year}{2019}).
\newblock


\bibitem[Krizhevsky(2014)]%
        {krizhevsky2014weird}
\bibfield{author}{\bibinfo{person}{Alex Krizhevsky}.} \bibinfo{year}{2014}\natexlab{}.
\newblock \showarticletitle{One weird trick for parallelizing convolutional neural networks}.
\newblock \bibinfo{journal}{\emph{ArXiv e-prints}} (\bibinfo{year}{2014}).
\newblock


\bibitem[LeCun et~al\mbox{.}(2015)]%
        {lecun2015deep}
\bibfield{author}{\bibinfo{person}{Yann LeCun}, \bibinfo{person}{Yoshua Bengio}, {and} \bibinfo{person}{Geoffrey Hinton}.} \bibinfo{year}{2015}\natexlab{}.
\newblock \showarticletitle{Deep learning}.
\newblock \bibinfo{journal}{\emph{Nature}} \bibinfo{volume}{521}, \bibinfo{number}{7553} (\bibinfo{year}{2015}), \bibinfo{pages}{436--444}.
\newblock


\bibitem[Lee et~al\mbox{.}(2018)]%
        {lee2018self}
\bibfield{author}{\bibinfo{person}{Seung~Hyun Lee}, \bibinfo{person}{Dae~Ha Kim}, {and} \bibinfo{person}{Byung~Cheol Song}.} \bibinfo{year}{2018}\natexlab{}.
\newblock \showarticletitle{Self-supervised knowledge distillation using singular value decomposition}. In \bibinfo{booktitle}{\emph{Proceedings of European Conference on Computer Vision (ECCV)}}.
\newblock


\bibitem[Lee et~al\mbox{.}(2019)]%
        {lee2019defending}
\bibfield{author}{\bibinfo{person}{Taesung Lee}, \bibinfo{person}{Benjamin Edwards}, \bibinfo{person}{Ian Molloy}, {and} \bibinfo{person}{Dong Su}.} \bibinfo{year}{2019}\natexlab{}.
\newblock \showarticletitle{Defending against neural network model stealing attacks using deceptive perturbations}. In \bibinfo{booktitle}{\emph{2019 IEEE Security and Privacy Workshops (SPW)}}. IEEE, \bibinfo{pages}{43--49}.
\newblock


\bibitem[Li et~al\mbox{.}(2021)]%
        {li2021towards}
\bibfield{author}{\bibinfo{person}{Changjiang Li}, \bibinfo{person}{Shouling Ji}, \bibinfo{person}{Haiqin Weng}, \bibinfo{person}{Bo Li}, \bibinfo{person}{Jie Shi}, \bibinfo{person}{Raheem Beyah}, \bibinfo{person}{Shanqing Guo}, \bibinfo{person}{Zonghui Wang}, {and} \bibinfo{person}{Ting Wang}.} \bibinfo{year}{2021}\natexlab{}.
\newblock \showarticletitle{Towards certifying the asymmetric robustness for neural networks: quantification and applications}.
\newblock \bibinfo{journal}{\emph{IEEE Transactions on Dependable and Secure Computing}} \bibinfo{volume}{19}, \bibinfo{number}{6} (\bibinfo{year}{2021}), \bibinfo{pages}{3987--4001}.
\newblock


\bibitem[Li et~al\mbox{.}(2023)]%
        {li2023embarrassingly}
\bibfield{author}{\bibinfo{person}{Changjiang Li}, \bibinfo{person}{Ren Pang}, \bibinfo{person}{Zhaohan Xi}, \bibinfo{person}{Tianyu Du}, \bibinfo{person}{Shouling Ji}, \bibinfo{person}{Yuan Yao}, {and} \bibinfo{person}{Ting Wang}.} \bibinfo{year}{2023}\natexlab{}.
\newblock \showarticletitle{An Embarrassingly Simple Backdoor Attack on Self-supervised Learning}. In \bibinfo{booktitle}{\emph{The 2023 International Conference on Computer Vision (ICCV' 23)}}.
\newblock


\bibitem[Li et~al\mbox{.}(2022)]%
        {li2022seeing}
\bibfield{author}{\bibinfo{person}{Changjiang Li}, \bibinfo{person}{Li Wang}, \bibinfo{person}{Shouling Ji}, \bibinfo{person}{Xuhong Zhang}, \bibinfo{person}{Zhaohan Xi}, \bibinfo{person}{Shanqing Guo}, {and} \bibinfo{person}{Ting Wang}.} \bibinfo{year}{2022}\natexlab{}.
\newblock \showarticletitle{Seeing is living? rethinking the security of facial liveness verification in the deepfake era}.
\newblock \bibinfo{journal}{\emph{USENIX Security 2022}} (\bibinfo{year}{2022}).
\newblock


\bibitem[Li et~al\mbox{.}(2019)]%
        {li2019det}
\bibfield{author}{\bibinfo{person}{Changjiang Li}, \bibinfo{person}{Haiqin Weng}, \bibinfo{person}{Shouling Ji}, \bibinfo{person}{Jianfeng Dong}, {and} \bibinfo{person}{Qinming He}.} \bibinfo{year}{2019}\natexlab{}.
\newblock \showarticletitle{DeT: Defending against adversarial examples via decreasing transferability}. In \bibinfo{booktitle}{\emph{Cyberspace Safety and Security: 11th International Symposium, CSS 2019, Guangzhou, China, December 1--3, 2019, Proceedings, Part I 11}}. Springer International Publishing, \bibinfo{pages}{307--322}.
\newblock


\bibitem[Li and Deng(2019)]%
        {li2019reliable}
\bibfield{author}{\bibinfo{person}{Shan Li} {and} \bibinfo{person}{Weihong Deng}.} \bibinfo{year}{2019}\natexlab{}.
\newblock \showarticletitle{Reliable Crowdsourcing and Deep Locality-Preserving Learning for Unconstrained Facial Expression Recognition}.
\newblock \bibinfo{journal}{\emph{IEEE Transactions on Image Processing}} \bibinfo{volume}{28}, \bibinfo{number}{1} (\bibinfo{year}{2019}), \bibinfo{pages}{356--370}.
\newblock


\bibitem[Li et~al\mbox{.}(2024)]%
        {li2024feature}
\bibfield{author}{\bibinfo{person}{Zhenglin Li}, \bibinfo{person}{Yangchen Huang}, \bibinfo{person}{Mengran Zhu}, \bibinfo{person}{Jingyu Zhang}, \bibinfo{person}{JingHao Chang}, {and} \bibinfo{person}{Houze Liu}.} \bibinfo{year}{2024}\natexlab{}.
\newblock \showarticletitle{Feature Manipulation for DDPM based Change Detection}.
\newblock \bibinfo{journal}{\emph{arXiv preprint arXiv:2403.15943}} (\bibinfo{year}{2024}).
\newblock


\bibitem[Liang et~al\mbox{.}(2021)]%
        {liang2021omnilytics}
\bibfield{author}{\bibinfo{person}{Jiacheng Liang}, \bibinfo{person}{Songze Li}, \bibinfo{person}{Bochuan Cao}, \bibinfo{person}{Wensi Jiang}, {and} \bibinfo{person}{Chaoyang He}.} \bibinfo{year}{2021}\natexlab{}.
\newblock \showarticletitle{Omnilytics: A blockchain-based secure data market for decentralized machine learning}.
\newblock \bibinfo{journal}{\emph{arXiv preprint arXiv:2107.05252}} (\bibinfo{year}{2021}).
\newblock


\bibitem[Liu et~al\mbox{.}(2023b)]%
        {liu2023slowlidar}
\bibfield{author}{\bibinfo{person}{Han Liu}, \bibinfo{person}{Yuhao Wu}, \bibinfo{person}{Zhiyuan Yu}, \bibinfo{person}{Yevgeniy Vorobeychik}, {and} \bibinfo{person}{Ning Zhang}.} \bibinfo{year}{2023}\natexlab{b}.
\newblock \showarticletitle{Slowlidar: Increasing the latency of lidar-based detection using adversarial examples}. In \bibinfo{booktitle}{\emph{Proceedings of the IEEE/CVF Conference on Computer Vision and Pattern Recognition}}. \bibinfo{pages}{5146--5155}.
\newblock


\bibitem[Liu et~al\mbox{.}(2024)]%
        {liu2024please}
\bibfield{author}{\bibinfo{person}{Han Liu}, \bibinfo{person}{Yuhao Wu}, \bibinfo{person}{Zhiyuan Yu}, {and} \bibinfo{person}{Ning Zhang}.} \bibinfo{year}{2024}\natexlab{}.
\newblock \showarticletitle{Please Tell Me More: Privacy Impact of Explainability through the Lens of Membership Inference Attack}. In \bibinfo{booktitle}{\emph{2024 IEEE Symposium on Security and Privacy (SP)}}. IEEE Computer Society, \bibinfo{pages}{120--120}.
\newblock


\bibitem[Liu et~al\mbox{.}(2019)]%
        {liu2019roberta}
\bibfield{author}{\bibinfo{person}{Yinhan Liu}, \bibinfo{person}{Myle Ott}, \bibinfo{person}{Naman Goyal}, \bibinfo{person}{Jingfei Du}, \bibinfo{person}{Mandar Joshi}, \bibinfo{person}{Danqi Chen}, \bibinfo{person}{Omer Levy}, \bibinfo{person}{Mike Lewis}, \bibinfo{person}{Luke Zettlemoyer}, {and} \bibinfo{person}{Veselin Stoyanov}.} \bibinfo{year}{2019}\natexlab{}.
\newblock \showarticletitle{Roberta: A robustly optimized bert pretraining approach}.
\newblock \bibinfo{journal}{\emph{ArXiv e-prints}} (\bibinfo{year}{2019}).
\newblock


\bibitem[Liu et~al\mbox{.}(2022)]%
        {liu2022efficient}
\bibfield{author}{\bibinfo{person}{Ziyao Liu}, \bibinfo{person}{Jiale Guo}, \bibinfo{person}{Kwok-Yan Lam}, {and} \bibinfo{person}{Jun Zhao}.} \bibinfo{year}{2022}\natexlab{}.
\newblock \showarticletitle{Efficient dropout-resilient aggregation for privacy-preserving machine learning}.
\newblock \bibinfo{journal}{\emph{IEEE Transactions on Information Forensics and Security}}  \bibinfo{volume}{18} (\bibinfo{year}{2022}), \bibinfo{pages}{1839--1854}.
\newblock


\bibitem[Liu et~al\mbox{.}(2023a)]%
        {liu2023long}
\bibfield{author}{\bibinfo{person}{Ziyao Liu}, \bibinfo{person}{Hsiao-Ying Lin}, {and} \bibinfo{person}{Yamin Liu}.} \bibinfo{year}{2023}\natexlab{a}.
\newblock \showarticletitle{Long-term privacy-preserving aggregation with user-dynamics for federated learning}.
\newblock \bibinfo{journal}{\emph{IEEE Transactions on Information Forensics and Security}} (\bibinfo{year}{2023}).
\newblock


\bibitem[Loshchilov and Hutter(2019)]%
        {adamw}
\bibfield{author}{\bibinfo{person}{Ilya Loshchilov} {and} \bibinfo{person}{Frank Hutter}.} \bibinfo{year}{2019}\natexlab{}.
\newblock \showarticletitle{Decoupled Weight Decay Regularization}.
\newblock \bibinfo{journal}{\emph{ArXiv e-prints}} (\bibinfo{year}{2019}).
\newblock


\bibitem[Lundqvist et~al\mbox{.}(2022)]%
        {lundqvist2022karolinska}
\bibfield{author}{\bibinfo{person}{Daniel Lundqvist}, \bibinfo{person}{Anders Flykt}, {and} \bibinfo{person}{A {\"O}hman}.} \bibinfo{year}{2022}\natexlab{}.
\newblock \showarticletitle{The Karolinska directed emotional faces—KDEF, CD ROM from Department of Clinical Neuroscience, Psychology section, Karolinska Institutet, 1998}.
\newblock \bibinfo{journal}{\emph{ArXiv e-prints}} (\bibinfo{year}{2022}).
\newblock


\bibitem[Lyu et~al\mbox{.}(2024)]%
        {lyu2024task}
\bibfield{author}{\bibinfo{person}{Weimin Lyu}, \bibinfo{person}{Xiao Lin}, \bibinfo{person}{Songzhu Zheng}, \bibinfo{person}{Lu Pang}, \bibinfo{person}{Haibin Ling}, \bibinfo{person}{Susmit Jha}, {and} \bibinfo{person}{Chao Chen}.} \bibinfo{year}{2024}\natexlab{}.
\newblock \showarticletitle{Task-Agnostic Detector for Insertion-Based Backdoor Attacks}.
\newblock \bibinfo{journal}{\emph{arXiv preprint arXiv:2403.17155}} (\bibinfo{year}{2024}).
\newblock


\bibitem[Lyu et~al\mbox{.}(2023)]%
        {lyu2023attention}
\bibfield{author}{\bibinfo{person}{Weimin Lyu}, \bibinfo{person}{Songzhu Zheng}, \bibinfo{person}{Lu Pang}, \bibinfo{person}{Haibin Ling}, {and} \bibinfo{person}{Chao Chen}.} \bibinfo{year}{2023}\natexlab{}.
\newblock \showarticletitle{Attention-Enhancing Backdoor Attacks Against BERT-based Models}.
\newblock \bibinfo{journal}{\emph{arXiv preprint arXiv:2310.14480}} (\bibinfo{year}{2023}).
\newblock


\bibitem[Madry et~al\mbox{.}(2018)]%
        {madry2018towards}
\bibfield{author}{\bibinfo{person}{Aleksander Madry}, \bibinfo{person}{Aleksandar Makelov}, \bibinfo{person}{Ludwig Schmidt}, \bibinfo{person}{Dimitris Tsipras}, {and} \bibinfo{person}{Adrian Vladu}.} \bibinfo{year}{2018}\natexlab{}.
\newblock \showarticletitle{Towards deep learning models resistant to adversarial attacks}. In \bibinfo{booktitle}{\emph{Proceedings of International Conference on Learning Representations (ICLR)}}.
\newblock


\bibitem[Mittal et~al\mbox{.}(2019)]%
        {mittal2019parting}
\bibfield{author}{\bibinfo{person}{Sudhanshu Mittal}, \bibinfo{person}{Maxim Tatarchenko}, \bibinfo{person}{{\"O}zg{\"u}n {\c{C}}i{\c{c}}ek}, {and} \bibinfo{person}{Thomas Brox}.} \bibinfo{year}{2019}\natexlab{}.
\newblock \showarticletitle{Parting with illusions about deep active learning}.
\newblock \bibinfo{journal}{\emph{ArXiv e-prints}} (\bibinfo{year}{2019}).
\newblock


\bibitem[Oh et~al\mbox{.}(2019)]%
        {oh2019towards}
\bibfield{author}{\bibinfo{person}{Seong~Joon Oh}, \bibinfo{person}{Bernt Schiele}, {and} \bibinfo{person}{Mario Fritz}.} \bibinfo{year}{2019}\natexlab{}.
\newblock \showarticletitle{Towards reverse-engineering black-box neural networks}.
\newblock \bibinfo{journal}{\emph{Explainable AI: Interpreting, Explaining and Visualizing Deep Learning}} (\bibinfo{year}{2019}), \bibinfo{pages}{121--144}.
\newblock


\bibitem[Orekondy et~al\mbox{.}(2019a)]%
        {orekondy2019knockoff}
\bibfield{author}{\bibinfo{person}{Tribhuvanesh Orekondy}, \bibinfo{person}{Bernt Schiele}, {and} \bibinfo{person}{Mario Fritz}.} \bibinfo{year}{2019}\natexlab{a}.
\newblock \showarticletitle{Knockoff nets: Stealing functionality of black-box models}. In \bibinfo{booktitle}{\emph{Proceedings of IEEE Conference on Computer Vision and Pattern Recognition (CVPR)}}.
\newblock


\bibitem[Orekondy et~al\mbox{.}(2019b)]%
        {orekondy2019prediction}
\bibfield{author}{\bibinfo{person}{Tribhuvanesh Orekondy}, \bibinfo{person}{Bernt Schiele}, {and} \bibinfo{person}{Mario Fritz}.} \bibinfo{year}{2019}\natexlab{b}.
\newblock \showarticletitle{Prediction poisoning: Towards defenses against dnn model stealing attacks}.
\newblock \bibinfo{journal}{\emph{arXiv preprint arXiv:1906.10908}} (\bibinfo{year}{2019}).
\newblock


\bibitem[Pal et~al\mbox{.}(2019)]%
        {pal2019framework}
\bibfield{author}{\bibinfo{person}{Soham Pal}, \bibinfo{person}{Yash Gupta}, \bibinfo{person}{Aditya Shukla}, \bibinfo{person}{Aditya Kanade}, \bibinfo{person}{Shirish Shevade}, {and} \bibinfo{person}{Vinod Ganapathy}.} \bibinfo{year}{2019}\natexlab{}.
\newblock \showarticletitle{A framework for the extraction of deep neural networks by leveraging public data}.
\newblock \bibinfo{journal}{\emph{ArXiv e-prints}} (\bibinfo{year}{2019}).
\newblock


\bibitem[Pal et~al\mbox{.}(2020)]%
        {pal2020activethief}
\bibfield{author}{\bibinfo{person}{Soham Pal}, \bibinfo{person}{Yash Gupta}, \bibinfo{person}{Aditya Shukla}, \bibinfo{person}{Aditya Kanade}, \bibinfo{person}{Shirish Shevade}, {and} \bibinfo{person}{Vinod Ganapathy}.} \bibinfo{year}{2020}\natexlab{}.
\newblock \showarticletitle{Activethief: Model extraction using active learning and unannotated public data}. In \bibinfo{booktitle}{\emph{Proceedings of AAAI Conference on Artificial Intelligence (AAAI)}}.
\newblock


\bibitem[Papernot et~al\mbox{.}(2017)]%
        {papernot2017practical}
\bibfield{author}{\bibinfo{person}{Nicolas Papernot}, \bibinfo{person}{Patrick McDaniel}, \bibinfo{person}{Ian Goodfellow}, \bibinfo{person}{Somesh Jha}, \bibinfo{person}{Z~Berkay Celik}, {and} \bibinfo{person}{Ananthram Swami}.} \bibinfo{year}{2017}\natexlab{}.
\newblock \showarticletitle{Practical black-box attacks against machine learning}. In \bibinfo{booktitle}{\emph{Proceedings of ACM Symposium on Information, Computer and Communications Security (AsiaCCS)}}.
\newblock


\bibitem[Papernot et~al\mbox{.}(2018)]%
        {papernot2018sok}
\bibfield{author}{\bibinfo{person}{Nicolas Papernot}, \bibinfo{person}{Patrick McDaniel}, \bibinfo{person}{Arunesh Sinha}, {and} \bibinfo{person}{Michael~P Wellman}.} \bibinfo{year}{2018}\natexlab{}.
\newblock \showarticletitle{Sok: Security and privacy in machine learning}. In \bibinfo{booktitle}{\emph{Proceedings of IEEE European Symposium on Security and Privacy (Euro S\&P)}}.
\newblock


\bibitem[Papernot et~al\mbox{.}(2016b)]%
        {papernot2016distillation}
\bibfield{author}{\bibinfo{person}{Nicolas Papernot}, \bibinfo{person}{Patrick McDaniel}, \bibinfo{person}{Xi Wu}, \bibinfo{person}{Somesh Jha}, {and} \bibinfo{person}{Ananthram Swami}.} \bibinfo{year}{2016}\natexlab{b}.
\newblock \showarticletitle{Distillation as a defense to adversarial perturbations against deep neural networks}. In \bibinfo{booktitle}{\emph{Proceedings of IEEE Symposium on Security and Privacy (S\&P)}}.
\newblock


\bibitem[Papernot et~al\mbox{.}(2016a)]%
        {papernot2016transferability}
\bibfield{author}{\bibinfo{person}{Nicolas Papernot}, \bibinfo{person}{Patrick~D McDaniel}, {and} \bibinfo{person}{Ian~J Goodfellow}.} \bibinfo{year}{2016}\natexlab{a}.
\newblock \showarticletitle{Transferability in machine learning: from phenomena to black-box attacks using adversarial samples}.
\newblock \bibinfo{journal}{\emph{ArXiv e-prints}} (\bibinfo{year}{2016}).
\newblock


\bibitem[Pengcheng et~al\mbox{.}(2018)]%
        {pengcheng2018query}
\bibfield{author}{\bibinfo{person}{Li Pengcheng}, \bibinfo{person}{Jinfeng Yi}, {and} \bibinfo{person}{Lijun Zhang}.} \bibinfo{year}{2018}\natexlab{}.
\newblock \showarticletitle{Query-efficient black-box attack by active learning}. In \bibinfo{booktitle}{\emph{Proceedings of IEEE International Conference on Data Mining (ICDM)}}.
\newblock


\bibitem[Qi et~al\mbox{.}(2023)]%
        {qi2023ocbev}
\bibfield{author}{\bibinfo{person}{Zhangyang Qi}, \bibinfo{person}{Jiaqi Wang}, \bibinfo{person}{Xiaoyang Wu}, {and} \bibinfo{person}{Hengshuang Zhao}.} \bibinfo{year}{2023}\natexlab{}.
\newblock \showarticletitle{OCBEV: Object-Centric BEV Transformer for Multi-View 3D Object Detection}.
\newblock \bibinfo{journal}{\emph{arXiv preprint arXiv:2306.01738}} (\bibinfo{year}{2023}).
\newblock


\bibitem[Robbins and Monro(1951)]%
        {sgd}
\bibfield{author}{\bibinfo{person}{Herbert Robbins} {and} \bibinfo{person}{Sutton Monro}.} \bibinfo{year}{1951}\natexlab{}.
\newblock \showarticletitle{A stochastic approximation method}.
\newblock \bibinfo{journal}{\emph{The Annals of Mathematical Statistics}} (\bibinfo{year}{1951}), \bibinfo{pages}{400--407}.
\newblock


\bibitem[Romero et~al\mbox{.}(2014)]%
        {romero2014fitnets}
\bibfield{author}{\bibinfo{person}{Adriana Romero}, \bibinfo{person}{Nicolas Ballas}, \bibinfo{person}{Samira~Ebrahimi Kahou}, \bibinfo{person}{Antoine Chassang}, \bibinfo{person}{Carlo Gatta}, {and} \bibinfo{person}{Yoshua Bengio}.} \bibinfo{year}{2014}\natexlab{}.
\newblock \showarticletitle{Fitnets: Hints for thin deep nets}.
\newblock \bibinfo{journal}{\emph{ArXiv e-prints}} (\bibinfo{year}{2014}).
\newblock


\bibitem[Sener and Savarese(2017)]%
        {sener2017active}
\bibfield{author}{\bibinfo{person}{Ozan Sener} {and} \bibinfo{person}{Silvio Savarese}.} \bibinfo{year}{2017}\natexlab{}.
\newblock \showarticletitle{Active learning for convolutional neural networks: A core-set approach}.
\newblock \bibinfo{journal}{\emph{ArXiv e-prints}} (\bibinfo{year}{2017}).
\newblock


\bibitem[Shi et~al\mbox{.}(2017)]%
        {shi2017steal}
\bibfield{author}{\bibinfo{person}{Yi Shi}, \bibinfo{person}{Yalin Sagduyu}, {and} \bibinfo{person}{Alexander Grushin}.} \bibinfo{year}{2017}\natexlab{}.
\newblock \showarticletitle{How to steal a machine learning classifier with deep learning}. In \bibinfo{booktitle}{\emph{Proceedings of the IEEE International Symposium on Technologies for Homeland Security (HST)}}.
\newblock


\bibitem[Shi et~al\mbox{.}(2018)]%
        {shi2018active}
\bibfield{author}{\bibinfo{person}{Yi Shi}, \bibinfo{person}{Yalin~E Sagduyu}, \bibinfo{person}{Kemal Davaslioglu}, {and} \bibinfo{person}{Jason~H Li}.} \bibinfo{year}{2018}\natexlab{}.
\newblock \showarticletitle{Active deep learning attacks under strict rate limitations for online API calls}. In \bibinfo{booktitle}{\emph{2018 IEEE International Symposium on Technologies for Homeland Security (HST)}}.
\newblock


\bibitem[Simonyan and Zisserman(2014)]%
        {simonyan2014very}
\bibfield{author}{\bibinfo{person}{Karen Simonyan} {and} \bibinfo{person}{Andrew Zisserman}.} \bibinfo{year}{2014}\natexlab{}.
\newblock \showarticletitle{Very deep convolutional networks for large-scale image recognition}.
\newblock \bibinfo{journal}{\emph{ArXiv e-prints}} (\bibinfo{year}{2014}).
\newblock


\bibitem[Szegedy et~al\mbox{.}(2014)]%
        {lenet}
\bibfield{author}{\bibinfo{person}{Christian Szegedy}, \bibinfo{person}{Wei Liu}, \bibinfo{person}{Yangqing Jia}, \bibinfo{person}{Pierre Sermanet}, \bibinfo{person}{Scott~E. Reed}, \bibinfo{person}{Dragomir Anguelov}, \bibinfo{person}{Dumitru Erhan}, \bibinfo{person}{Vincent Vanhoucke}, {and} \bibinfo{person}{Andrew Rabinovich}.} \bibinfo{year}{2014}\natexlab{}.
\newblock \showarticletitle{Going Deeper with Convolutions}.
\newblock \bibinfo{journal}{\emph{ArXiv e-prints}} (\bibinfo{year}{2014}).
\newblock


\bibitem[Tan and Le(2020)]%
        {tan2020efficientnet}
\bibfield{author}{\bibinfo{person}{Mingxing Tan} {and} \bibinfo{person}{Quoc~V. Le}.} \bibinfo{year}{2020}\natexlab{}.
\newblock \showarticletitle{EfficientNet: Rethinking Model Scaling for Convolutional Neural Networks}.
\newblock \bibinfo{journal}{\emph{ArXiv e-prints}} (\bibinfo{year}{2020}).
\newblock


\bibitem[Tram{\`e}r et~al\mbox{.}(2016)]%
        {tramer2016stealing}
\bibfield{author}{\bibinfo{person}{Florian Tram{\`e}r}, \bibinfo{person}{Fan Zhang}, \bibinfo{person}{Ari Juels}, \bibinfo{person}{Michael~K Reiter}, {and} \bibinfo{person}{Thomas Ristenpart}.} \bibinfo{year}{2016}\natexlab{}.
\newblock \showarticletitle{Stealing Machine Learning Models via Prediction APIs.}. In \bibinfo{booktitle}{\emph{Proceedings of USENIX Security Symposium (SEC)}}.
\newblock


\bibitem[Truong et~al\mbox{.}(2021)]%
        {Truong_2021_CVPR}
\bibfield{author}{\bibinfo{person}{Jean-Baptiste Truong}, \bibinfo{person}{Pratyush Maini}, \bibinfo{person}{Robert~J. Walls}, {and} \bibinfo{person}{Nicolas Papernot}.} \bibinfo{year}{2021}\natexlab{}.
\newblock \showarticletitle{Data-Free Model Extraction}. In \bibinfo{booktitle}{\emph{Proceedings of IEEE Conference on Computer Vision and Pattern Recognition (CVPR)}}.
\newblock


\bibitem[Wang and Gong(2019)]%
        {wang2019stealing}
\bibfield{author}{\bibinfo{person}{Binghui Wang} {and} \bibinfo{person}{Neil~Zhenqiang Gong}.} \bibinfo{year}{2019}\natexlab{}.
\newblock \showarticletitle{Stealing Hyperparameters in Machine Learning}.
\newblock \bibinfo{journal}{\emph{ArXiv e-prints}} (\bibinfo{year}{2019}).
\newblock


\bibitem[Wu et~al\mbox{.}(2021)]%
        {wu2021learning}
\bibfield{author}{\bibinfo{person}{Wei Wu}, \bibinfo{person}{Xiaoyuan Jing}, \bibinfo{person}{Wencai Du}, {and} \bibinfo{person}{Guoliang Chen}.} \bibinfo{year}{2021}\natexlab{}.
\newblock \showarticletitle{Learning dynamics of gradient descent optimization in deep neural networks}.
\newblock \bibinfo{journal}{\emph{Science China Information Sciences}}  \bibinfo{volume}{64} (\bibinfo{year}{2021}), \bibinfo{pages}{1--15}.
\newblock


\bibitem[Xu et~al\mbox{.}(2023)]%
        {xu2023toposemiseg}
\bibfield{author}{\bibinfo{person}{Meilong Xu}, \bibinfo{person}{Xiaoling Hu}, \bibinfo{person}{Saumya Gupta}, \bibinfo{person}{Shahira Abousamra}, {and} \bibinfo{person}{Chao Chen}.} \bibinfo{year}{2023}\natexlab{}.
\newblock \bibinfo{title}{TopoSemiSeg: Enforcing Topological Consistency for Semi-Supervised Segmentation of Histopathology Images}.
\newblock
\newblock
\showeprint[arxiv]{2311.16447}~[eess.IV]


\bibitem[Yang et~al\mbox{.}(2019)]%
        {yang2019xlnet}
\bibfield{author}{\bibinfo{person}{Zhilin Yang}, \bibinfo{person}{Zihang Dai}, \bibinfo{person}{Yiming Yang}, \bibinfo{person}{Jaime Carbonell}, \bibinfo{person}{Russ~R Salakhutdinov}, {and} \bibinfo{person}{Quoc~V Le}.} \bibinfo{year}{2019}\natexlab{}.
\newblock \showarticletitle{Xlnet: Generalized autoregressive pretraining for language understanding}. In \bibinfo{booktitle}{\emph{Proceedings of Advances in Neural Information Processing Systems (NeurIPS)}}, Vol.~\bibinfo{volume}{32}.
\newblock


\bibitem[Yim et~al\mbox{.}(2017)]%
        {yim2017gift}
\bibfield{author}{\bibinfo{person}{Junho Yim}, \bibinfo{person}{Donggyu Joo}, \bibinfo{person}{Jihoon Bae}, {and} \bibinfo{person}{Junmo Kim}.} \bibinfo{year}{2017}\natexlab{}.
\newblock \showarticletitle{A gift from knowledge distillation: Fast optimization, network minimization and transfer learning}. In \bibinfo{booktitle}{\emph{Proceedings of IEEE Conference on Computer Vision and Pattern Recognition (CVPR)}}.
\newblock


\bibitem[Yu et~al\mbox{.}(2020)]%
        {yu2020cloudleak}
\bibfield{author}{\bibinfo{person}{Honggang Yu}, \bibinfo{person}{Kaichen Yang}, \bibinfo{person}{Teng Zhang}, \bibinfo{person}{Yun-Yun Tsai}, \bibinfo{person}{Tsung-Yi Ho}, {and} \bibinfo{person}{Yier Jin}.} \bibinfo{year}{2020}\natexlab{}.
\newblock \showarticletitle{CloudLeak: Large-Scale Deep Learning Models Stealing Through Adversarial Examples}. In \bibinfo{booktitle}{\emph{NDSS}}.
\newblock


\bibitem[Zhang et~al\mbox{.}(2021)]%
        {zhang2021restore}
\bibfield{author}{\bibinfo{person}{Haichao Zhang}, \bibinfo{person}{Zhe-Ming Lu}, \bibinfo{person}{Hao Luo}, {and} \bibinfo{person}{Ya-Pei Feng}.} \bibinfo{year}{2021}\natexlab{}.
\newblock \showarticletitle{Restore DeepFakes video frames via identifying individual motion styles}.
\newblock \bibinfo{journal}{\emph{Electronics Letters}} \bibinfo{volume}{57}, \bibinfo{number}{4} (\bibinfo{year}{2021}), \bibinfo{pages}{183--186}.
\newblock


\bibitem[Zhang et~al\mbox{.}(2016)]%
        {SOCIALRELATION_2017}
\bibfield{author}{\bibinfo{person}{Zhanpeng Zhang}, \bibinfo{person}{Ping Luo}, \bibinfo{person}{Chen~Change Loy}, {and} \bibinfo{person}{Xiaoou Tang}.} \bibinfo{year}{2016}\natexlab{}.
\newblock \showarticletitle{From Facial Expression Recognition to Interpersonal Relation Prediction}.
\newblock \bibinfo{journal}{\emph{ArXiv e-prints}} (\bibinfo{year}{2016}).
\newblock


\bibitem[Zhou et~al\mbox{.}(2023)]%
        {zhou2023pass}
\bibfield{author}{\bibinfo{person}{Qihua Zhou}, \bibinfo{person}{Song Guo}, \bibinfo{person}{Jun Pan}, \bibinfo{person}{Jiacheng Liang}, \bibinfo{person}{Zhenda Xu}, {and} \bibinfo{person}{Jingren Zhou}.} \bibinfo{year}{2023}\natexlab{}.
\newblock \showarticletitle{PASS: Patch Automatic Skip Scheme for Efficient Real-Time Video Perception on Edge Devices}. In \bibinfo{booktitle}{\emph{Proceedings of AAAI Conference on Artificial Intelligence (AAAI)}}, Vol.~\bibinfo{volume}{37}.
\newblock


\end{thebibliography}
